\documentclass[10pt,twocolumn,letterpaper]{article}

\usepackage{cvpr}              

\usepackage{graphicx}
\usepackage{float} 
\usepackage[accsupp]{axessibility}

\definecolor{cvprblue}{rgb}{0.21,0.49,0.74}
\usepackage[pagebackref,breaklinks,colorlinks,allcolors=cvprblue]{hyperref}
\usepackage{subcaption}

\usepackage[accsupp]{axessibility}  

\def\paperID{ABAW-51} 
\def\confName{CVPR}
\def\confYear{2025}

\title{Read My Ears! Horse Ear Movement Detection for Equine Affective State Assessment}

\author{João Alves \\
Visual Analysis and Perception Lab, Aalborg University, Aalborg, Denmark \\
{\tt\small jmal@create.aau.dk}
\and
Pia Haubro Andersen \\
Department of Veterinary Clinical Sciences, University of Copenhagen, Copenhagen, Denmark\\
{\tt\small pia.haubro.andersen@slu.se}
\and
Rikke Gade\\
Visual Analysis and Perception Lab, Aalborg University, Aalborg, Denmark\\
{\tt\small rg@create.aau.dk}
}

\begin{document}
\maketitle
\begin{abstract}
The Equine Facial Action Coding System (EquiFACS) enables the systematic annotation of facial movements through distinct Action Units (AUs). It serves as a crucial tool for assessing affective states in horses by identifying subtle facial expressions associated with discomfort. However, the field of horse affective state assessment is constrained by the scarcity of annotated data, as manually labelling facial AUs is both time-consuming and costly. To address this challenge, automated annotation systems are essential for leveraging existing datasets and improving affective states detection tools.

In this work, we study different methods for specific ear AU detection and localization from horse videos. We leverage past works on deep learning-based video feature extraction combined with recurrent neural networks for the video classification task, as well as a classic optical flow based approach. We achieve 87.5\% classification accuracy of ear movement presence on a public horse video dataset, demonstrating the potential of our approach. We discuss future directions to develop these systems, with the aim of bridging the gap between automated AU detection and practical applications in equine welfare and veterinary diagnostics. Our code will be made publicly available at \href{https://github.com/jmalves5/read-my-ears}{https://github.com/jmalves5/read-my-ears}.
\end{abstract}    
\section{Introduction}

\begin{figure}
  \centering
  \includegraphics[width=\columnwidth]{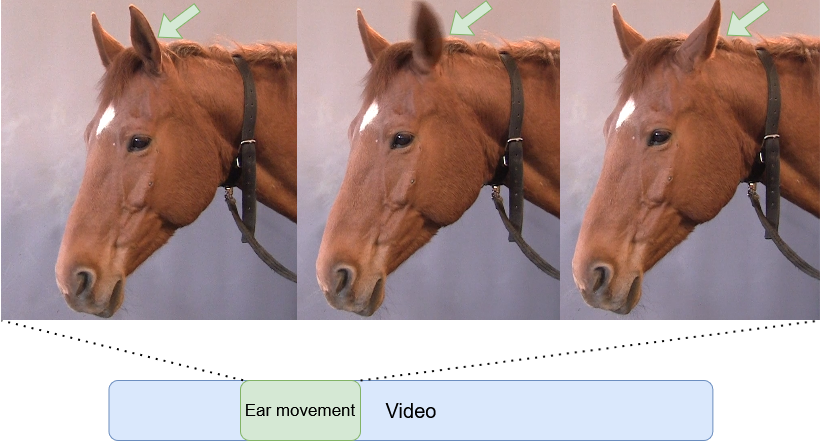}
  \caption{Ear rotator action unit (EAD104) example.}
  \label{fig:ear_rotation}
\end{figure}

Horses play an important role in multiple areas of our society, and thus, we have societal responsibility to ensure their well-being. Horses express pain through subtle facial movements that are often overlooked by the untrained human eye, potentially leading to late diagnoses. \cite{broomeSharingPainUsing2022}.

\begin{figure*}[t]
    \centering
    \begin{minipage}[t]{0.48\textwidth}
        \centering
        \includegraphics[width=0.8\linewidth]{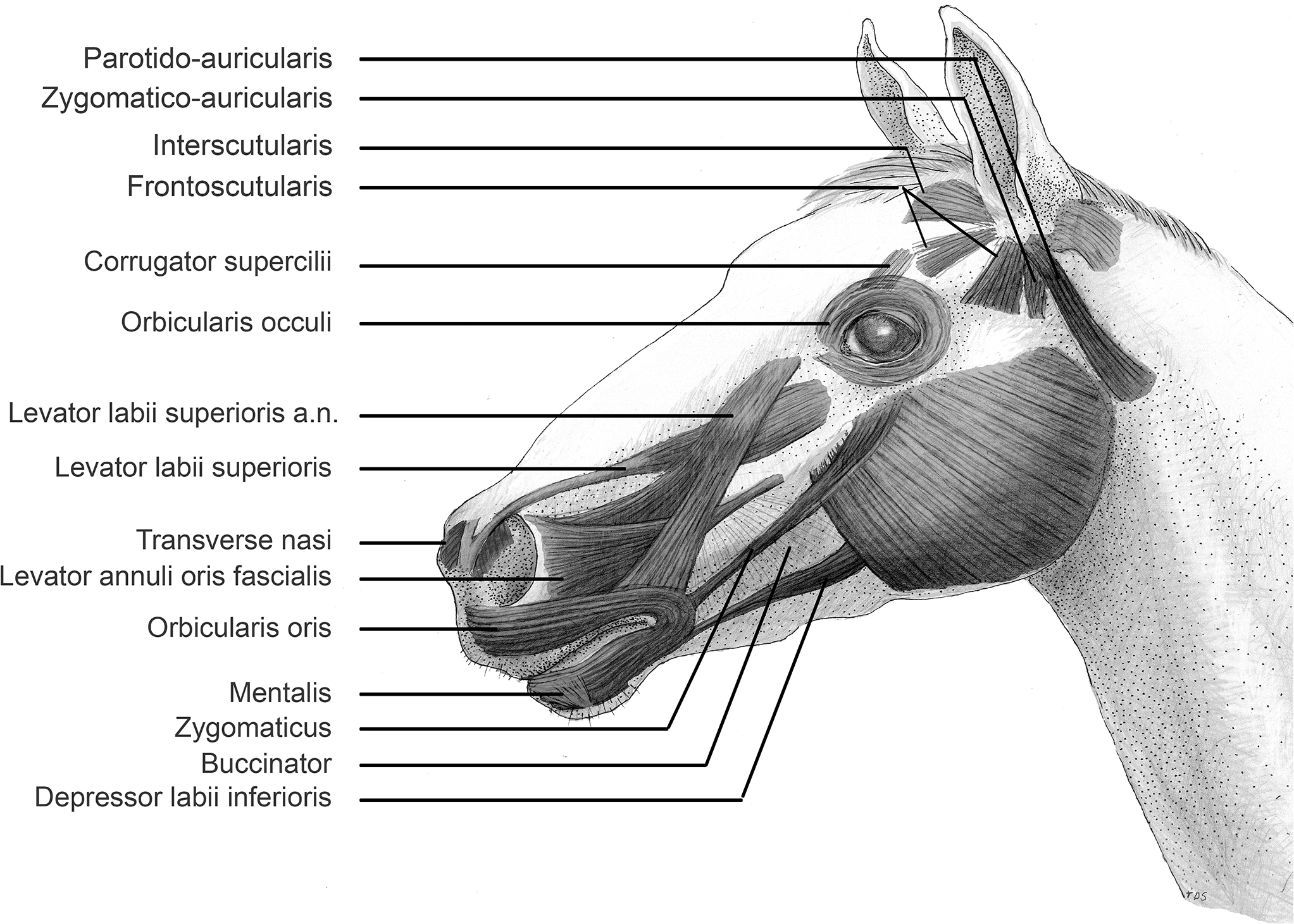}
        \caption{Horse facial muscles from \cite{wathanEquiFACSEquineFacial2015}.}
        \label{fig:facial_muscles}
    \end{minipage}
    \hfill
    \begin{minipage}[t]{0.48\textwidth}
        \centering
        \includegraphics[width=0.8\linewidth]{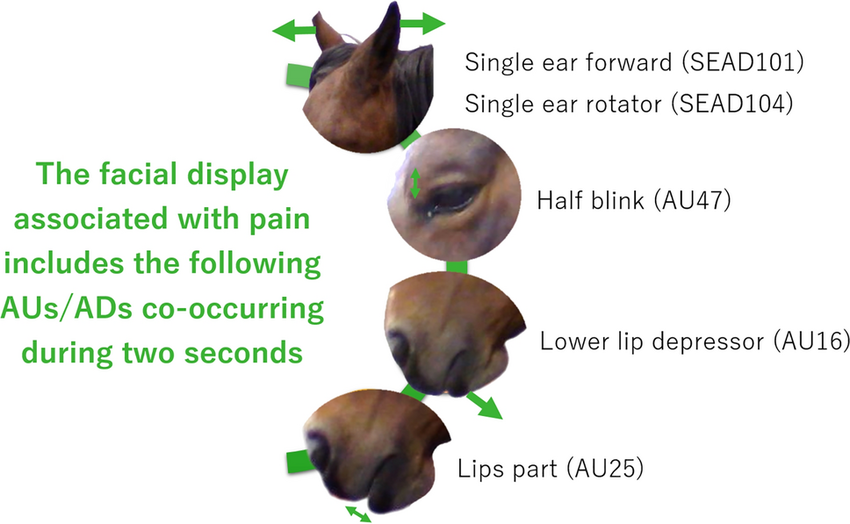}
        \caption{Example EquiFACS AUs from \cite{askChangesEquineFacial2024}.}
        \label{fig:exampleAUs}
    \end{minipage}
\end{figure*}

Literature on human pain provides several scales for pain assessment, but these usually rely heavily on self-reporting, except in cases where self-reporting is not possible \cite{broomeGoingDeeperTracking2023}. In such cases, an observational, objective analysis of physiological parameters is used instead. In particular, works focusing on facial expressions of pain have mostly used the Facial Action Coding System (FACS) \cite{ekman2002}, a coding system designed to describe human facial movements. As horses cannot communicate their feelings using language, observational scales such as the Horse Grimace Scale \cite{dallacostaDevelopmentHorseGrimace2014} and others are usually employed for pain assessment in these animals. Moreover, studies on facial pain responses in horses heavily rely on facial actions as defined by the Equine Facial Action Coding System (EquiFACS) \cite{wathanEquiFACSEquineFacial2015}, which allows for an objective evaluation of the animal’s facial movements, based on it's facial musculature (see Figure \ref{fig:facial_muscles} and Figure \ref{fig:exampleAUs}).

However, using EquiFACS presents significant practical challenges. The subtlety of horse facial movements necessitates that skilled veterinary workers manually annotate FACS in each video frame, a task that can take hours for even short video clips, making the annotation of horse data an extremely resource-demanding endeavour \cite{andersenMachineRecognitionFacial2021}. Moreover, Broome et al. \cite{broomeDynamicsAreImportant2019} highlighted the importance of facial movement dynamics in equine pain assessment, indicating the necessity of building systems that analyse horse video data rather than individual frames. This requirement further increases the complexity of annotation tasks. Currently, from a computer vision perspective, the biggest challenge in equine pain assessment is the lack of publicly available, large-scale, annotated video datasets \cite{broomeGoingDeeperTracking2023}. It is therefore important to study the automation of AU extraction to generate higher-quality datasets that can facilitate the development of improved horse pain assessment methods.
Past studies have investigated the relationship between specific facial movements and pain and studying their co-occurrence can provide valuable insights into the animal’s emotional state. In particular, ear-related AUs (see Figure \ref{fig:ear_rotation}) have been associated with equine affective states, like stress and pain \cite{dallacostaDevelopmentHorseGrimace2014, rashidEquineFacialAction2020, lundbladExploringFacialExpressions2024}. 

While detecting ear-related action units may initially appear straightforward, the subtlety and brevity of these movements, often accompanied by other head motions, present significant challenges in modelling and solving this problem. The scarcity of available public data further complicates the application of deep learning solutions.
With that in mind, this work focuses on the video clip classification of ear-related movements from horse video data with the goal of preforming action unit detection for horse affective state assessment. We propose and study different methodologies to automate the extraction of these movements and evaluate our methods on a publicly available dataset \cite{rashidEquineFacialAction2020}.

The main contributions of this work are as follows:
\begin{itemize}
    \item We propose a baseline approach and adapt two deep learning-based architectures (I3D+LSTM, VideoMAE+LSTM) for fine-grained equine AU identification.
    \item We demonstrate potential solutions to overcome the critical challenge of limited annotated data availability using data-efficient AU detection models.
    \item We take a step towards advancing animal welfare through the automated detection of key affective state indicators.
\end{itemize}

\section{Background and related works}
\subsection{EquiFACS for affective state assessment}
The Equine Facial Action Coding System (EquiFACS) provides a standardized coding system to objectively analyse and categorize horse facial expressions. At a high level, it systematically identifies and codes specific facial muscle movements, known as Action Units (AUs), that contribute to different horse facial expressions. This framework allows researchers and veterinarians to study equine affective states through a consistent methodology.

In \cite{rashidEquineFacialAction2020}, EquiFACS was applied to both experimental and clinical scenarios involving horses in pain to identify facial movements associated with acute, short-term pain. The study concluded that ear rotator movements, nostril dilation, and lower-face behaviours were important indicators of pain.

Similarly, in \cite{askChangesEquineFacial2024}, the authors examined how horses' facial expressions vary with the severity of orthopaedic pain. By using EquiFACS to objectively analyse facial movements in horses experiencing orthopaedic pain, the study found that AUs related to the ears, eyes, and lower-face regions (mouth, chin) were more prevalent during pain episodes. Additionally, the findings highlighted the importance of treating equine pain as a dynamic process, characterized by varying facial expressions over time.

In \cite{andersenMachineRecognitionFacial2021}, the authors explored the feasibility of an automated pain detection system for horses by integrating EquiFACS AU detection with machine learning techniques. To enhance keypoint detection accuracy, the authors applied cross-domain techniques by morphing animal features to human ones before feeding them into a standard model for human facial AU detection. Their approach yielded promising classification rates, demonstrating the potential for machine learning to advance automated equine pain recognition.

\subsection{Automated EquiFACS AU detection from videos}
Given the costly process of producing EquiFACS-annotated data \cite{andersenMachineRecognitionFacial2021}, this work aims to improve the annotation process for horse videos by studying methods for ear related AU detection. The overall goal of this work is to make annotation more cost-effective, thereby increasing both the number of available datasets and the quality of equine pain assessment tools built upon them.

Video action recognition is a fundamental computer vision task that identifies action instances performed in a video sequence. It is a well-established yet actively evolving topic in computer vision, with applications ranging from sports analysis, to scene understanding, to surveillance and beyond. The field has progressed from early handcrafted, feature-based approaches to modern deep-learning and transformer-based architectures, significantly improving accuracy and robustness.\par
With the rise of convolutional neural networks (CNNs) and recurrent neural networks (RNNs), significant improvements have been observed in methods leveraging the ability to learn spatiotemporal representations directly from video data \cite{simonyanTwoStreamConvolutionalNetworks2014, tranLearningSpatiotemporalFeatures2014, carreiraQuoVadisAction2017}. However, many of these models rely on frame-wise predictions rather than explicitly learning the duration of actions. This limitation paved the way for region proposal-based methods, which attempt to generate action segment candidates from video and then refine their temporal boundaries \cite{zhaoTemporalActionDetection2017, linBMNBoundaryMatchingNetwork2019, xuGTADSubGraphLocalization2020}.\par
The introduction of transformer models has significantly changed the field of action recognition, shifting away from handcrafted feature extraction and rigid proposal-based architectures toward more flexible models. In this context, transformer-based models have emerged as powerful feature extractors for action recognition, enhancing performance through rich, pre-trained spatiotemporal representations. VideoMAE \cite{tongVideoMAEMaskedAutoencoders2022} is a self-supervised learning model that extracts temporally contextualized features by reconstructing missing patches in video sequences.\par
Transformers specifically designed for action recognition, such as  RTD-Net \cite{tanRelaxedTransformerDecoders2021} and ActionFormer \cite{zhangActionFormerLocalizingMoments2022}, leverage hierarchical spatiotemporal attention to accurately detect action boundaries without relying on predefined proposals or anchor boxes. While these models improve localization accuracy and enhance generalization across different datasets and unseen actions, they require extensive domain-specific datasets for adaptation. Additionally, they struggle with capturing fine-grained boundaries, such as the subtle facial movements of horses.

Despite these advancements, applying video action recognition techniques to equine facial expression analysis presents unique challenges. Unlike human action recognition, where movements are often deliberate and easily distinguishable, equine facial movements—particularly ear-related AUs are subtle and may occur within short temporal windows. This necessitates models capable of fine-grained action recognition, ensuring that even brief and low-amplitude movements are accurately detected.

In this work, we focus on studying the feasibility of automated equine ear related AU detection, which have been linked to pain assessment. By leveraging both traditional motion analysis and deep learning-based approaches, we study action recognition methods and some of the specific challenges attached with the equine affective computing field. Our study evaluates the effectiveness of different methods in detecting these subtle ear movements, with the goal of advancing automated tools for equine welfare monitoring.
\section{Dataset}
\subsection{Dataset description}
\begin{figure*}
\centering
\includegraphics[width=0.85\linewidth]{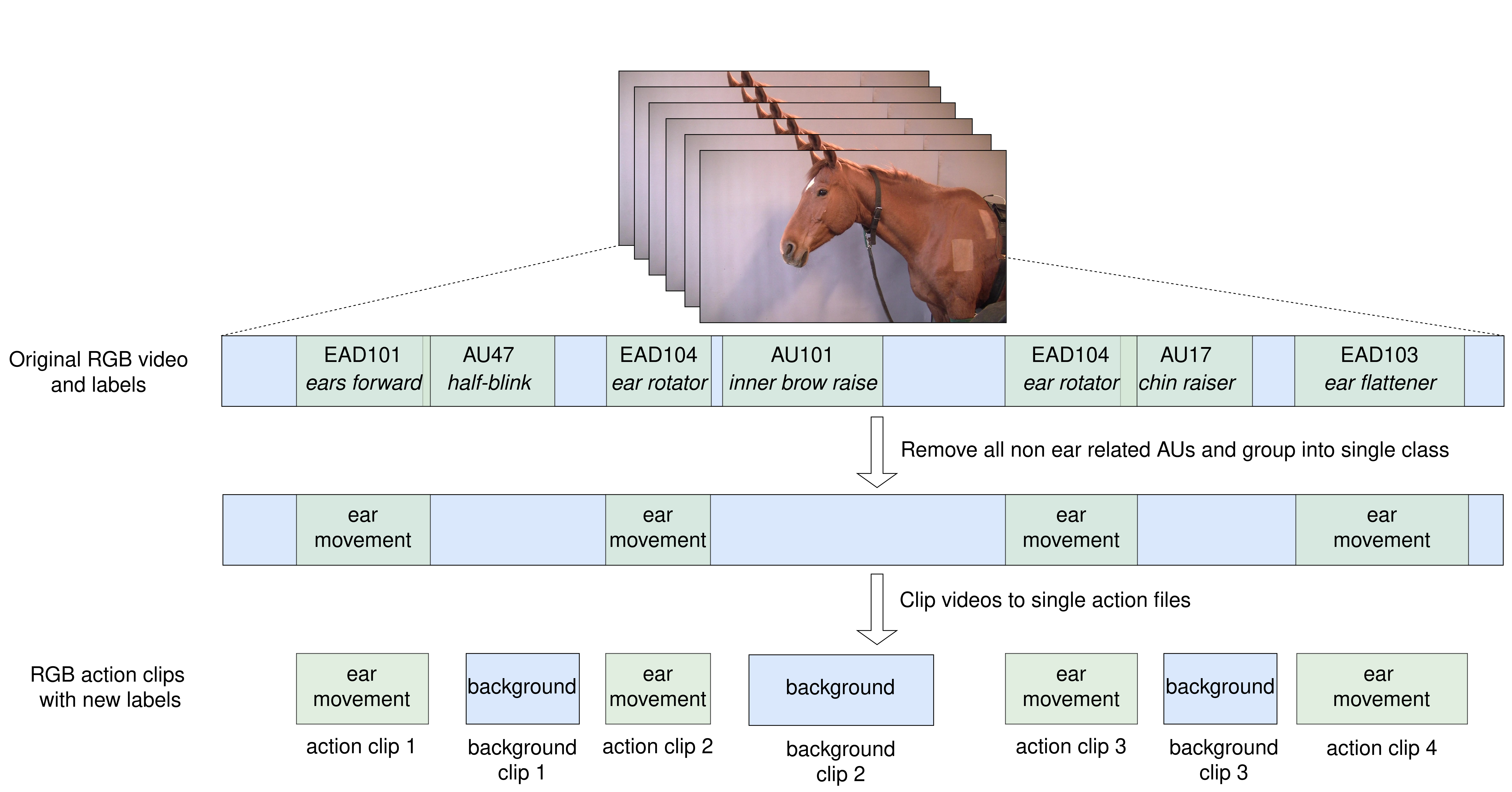}
\caption{Dataset processing from videos in \cite{rashidEquineFacialAction2020}.} \label{fig:tal_dataset}
\end{figure*}

For our experiments, we will use the data introduced in \cite{rashidEquineFacialAction2020}, which consists of 12 videos of horses (S1-S12) from different breeds recorded during a study on horses experiencing acute short-term pain\cite{gleerupEquinePainFace2015}. Each video setup consists of a static camera in a stable observing a horse subject that can freely move its head, including rotation, translation and, at times, going outside of the camera's field of view. This dataset contains expertly annotated EquiFACS labels for each of the 12 videos, providing a comprehensive resource for AU analysis (see Figure \ref{fig:tal_dataset}). In this work we adapt this data for the video classification task (see Section \ref{ssec:proprocess}) by clipping the relevant ear action sequences, as well as an close to equal number of background clips of similar lengths, ensuring a balanced dataset. Sample frames of the videos can be found in Figure \ref{fig:sample_frames} in the supplementary material.

\subsection{Pre-processing}\label{ssec:proprocess}
We prepare the dataset for binary video classification task using ear movement/no ear movement labels. First, we filter out all non ear related annotations, focusing exclusively on ear movements rather than other EquiFACS labels. Next, we extract clips from the original videos based on the remaining annotations, ensuring that each clip accurately captures the target action, ensuring that each segment contains only a single instance of ear movement. Then we clip background clips from the videos with random duration between 0.5 and 3 seconds. We make sure to extract a number of background clips that ensures a class balanced dataset (ear movement vs background). 

Moreover, to study the impact of frame rate on the action classification task, we employ RIFE \cite{huangRealTimeIntermediateFlow2020} to create videos with increased FPS of our video data via frame generation. Although quantitatively measuring the quality of the generated frames is challenging, we found the method's qualitative performance satisfactory and studied the method's performance on these videos (see Section \ref{experiments}).

Finally, data augmentation techniques were applied to increase the number of samples, using random horizontal flipping, as well as hue, brightness, saturation and contrast jittering.
\section{Methodology}

This section outlines the methodologies employed for detecting equine ear movements in video sequences, ranging from a classical optical flow-based method to advanced deep learning techniques. 

\subsection{Optical flow based ear movement detection baseline (movDet)}

As a baseline, we propose an optical flow based method that analyses the magnitude of optical flow vectors within a defined region of interest, in our case the horse’s ears. This approach provides a simple yet effective means of detecting movement by leveraging dense optical flow calculations.

\begin{figure*}
\centering
\includegraphics[width=\linewidth]{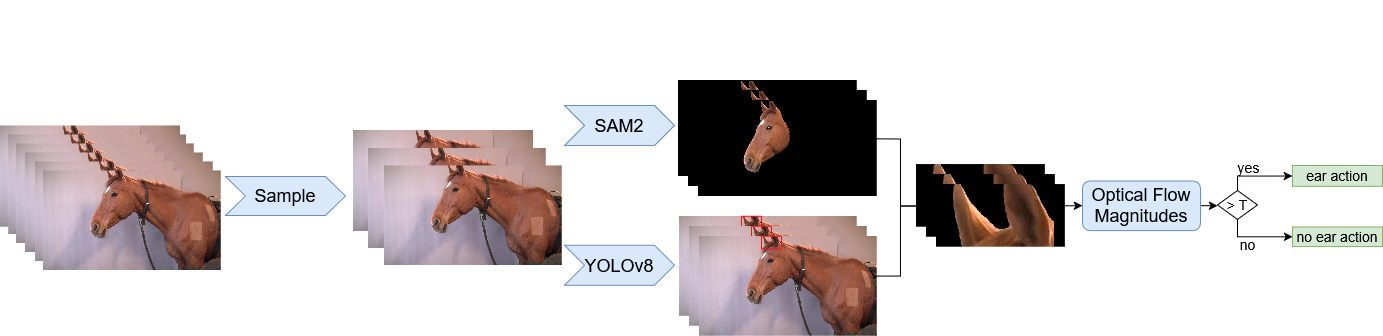}
\caption{Pipeline for the baseline optical flow based ear movement detection (movDet).} \label{fig:move_det}
\label{fig:movDet}
\end{figure*}

The process begins with background subtraction and ear detection to segment only the horse's ears. In order to do this, we train a YOLOv8 object detector specifically for ear detection using a custom dataset. To further refine segmentation, background subtraction is performed using SAM2 \cite{raviSAM2Segment2024}. We apply our method to the cropped detections of the horse's ears (see Figure \ref{fig:move_det}). 

Once the ears are detected, we compute Farnebäck's dense optical flow \cite{farnebackTwoFrameMotionEstimation2003} between consecutive cropped ear frames, sampled at a fixed rate throughout the video. The dense optical flow is then analysed across the entire clip, and the average magnitude of motion vectors is used to classify the presence of movement. A predefined threshold determines whether significant motion has occurred (see Figure \ref{fig:move_det}). This method serves as a benchmark for evaluating the effectiveness of more complex learning-based approaches.

\subsection{Inflated 3D ConvNet + LSTM (I3D+LSTM)} \label{i3d+lstm}
\begin{figure*}
\centering
\includegraphics[width=0.75\linewidth]{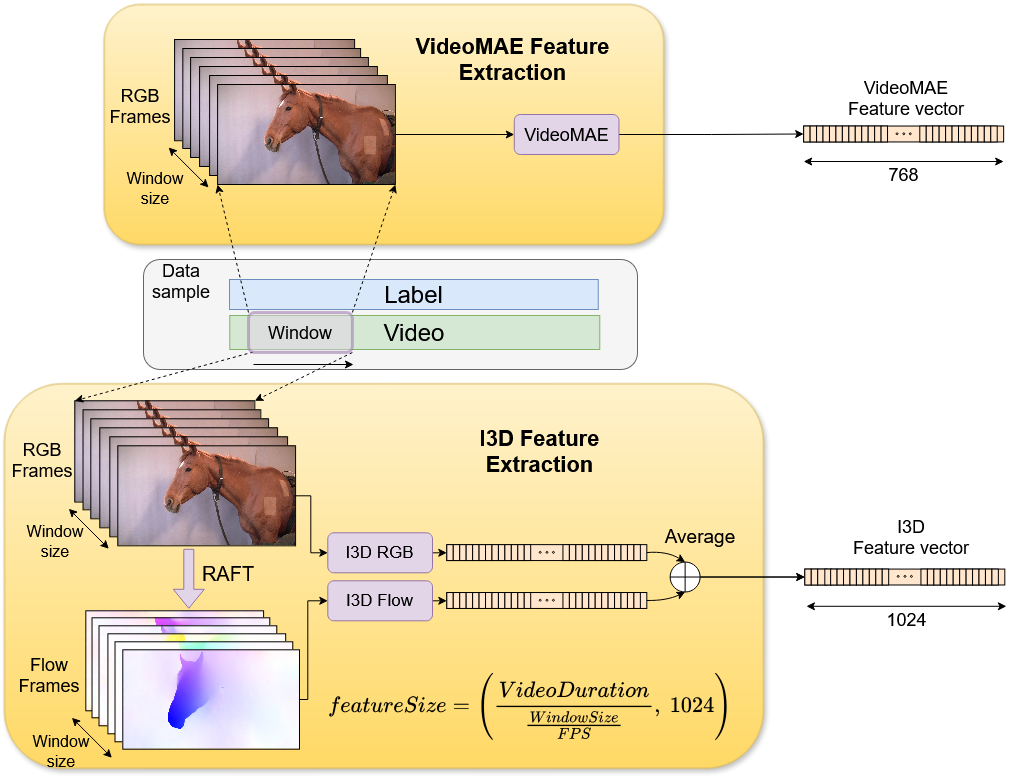}
\caption{Pipelines for feature extraction using VideoMAE and I3D methods. A mixed stream (RGB+Flow) late fusion approach is represented for the I3D case.} 
\label{fig:extraction}
\end{figure*}

Previous studies have demonstrated that convolutional neural networks (CNNs) trained on large-scale action recognition datasets can serve as effective feature extractors, particularly when incorporating temporal information. Notably, the Inflated 3D ConvNet (I3D) architecture introduced by \cite{carreiraQuoVadisAction2017} and associated Kinetics dataset, has shown strong performance in learning spatiotemporal representations. We adopt the approach proposed by \cite{wangI3DLSTMNewModel2019}, where features extracted using I3D are further processed by a Long Short-Term Memory (LSTM) network for binary classification, in our case of ear movement (action vs. no-action) for each video clip. We test our method using both RGB and optical flow streams (extracted via RAFT \cite{teedRAFTRecurrentAllPairs2020}) separately, as well as a late fusion strategy that averages each stream's feature vectors before feeding them into the LSTM. As the I3D model was originally trained on colour and flow streams, we test performance using colour, optical flow and mixed streams (using a late-fusion strategy, see Figure \ref{fig:extraction}).

LSTMs are well-suited for capturing long-term dependencies in sequential data, providing an advantage over I3D’s sliding window approach, which primarily encodes short-term dependencies through overlapping frame windows. Moreover, because LSTMs can process variable-length inputs, we can flexibly adjust I3D parameters, such as temporal window size and step size without needing to modify the network architecture.

To evaluate the effectiveness of our model, we experimented with LSTMs comprising two and three hidden layers, followed by a fully connected linear layer for binary classification before reaching the output layer using a sigmoid activation function. We tested hidden sizes of 256 and 512 neurons, applying a dropout rate of 0.2 for regularization. Training was conducted using Binary Cross-Entropy loss, with early stopping implemented based on a patience criterion of 20 epochs to prevent overfitting to our small dataset. 

\subsection{VideoMAE + LSTM}

\begin{figure*}
\centering
\includegraphics[width=0.7\linewidth]{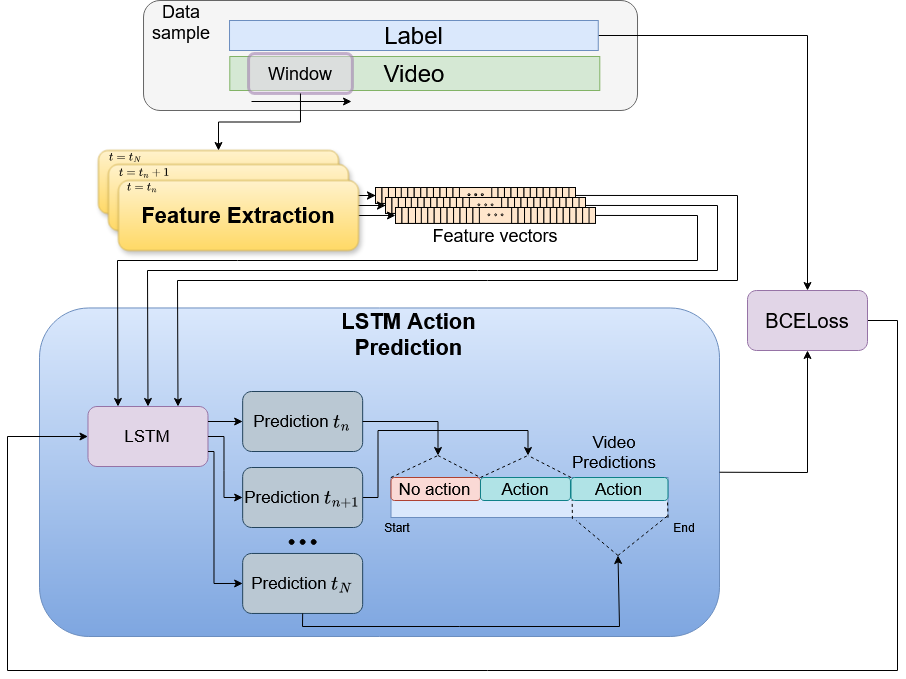}
\caption{LSTM architecture pipeline for movement detection method using I3D or VideoMAE features.} 
\label{fig:classification}
\end{figure*}

Video Masked AutoEncoder (VideoMAE) \cite{tongVideoMAEMaskedAutoencoders2022} has proven to be an efficient feature extractor for action recognition tasks. During training, VideoMAE samples frames to form 16-frame windows, which are then divided into spatiotemporal patches. A high masking ratio (e.g., 90\%) randomly hides most patches, and only the visible ones are processed by a Vision Transformer (ViT) to extract features. A lightweight decoder then reconstructs the missing patches, forcing the model to learn strong spatiotemporal representations through self-supervised learning. Several state-of-the-art methods utilize VideoMAE features as their input \cite{liuFineActionFineGrainedVideo2021}. Building on this idea, as well as the work in \cite{wangI3DLSTMNewModel2019}, we propose replacing the I3D feature extractor with a VideoMAE (ViT-B) model pre-trained and finetuned on the same Kinetics-400 dataset \cite{carreiraQuoVadisAction2017} (see Figure \ref{fig:extraction}). In this case we test the methods performance using the colour stream as VideoMAE was trained on colour information only.
A vector of size 768 is extracted from 16-frame windows after global spatiotemporal pooling, which then feed the same LSTM architecture (see Figure \ref{fig:extraction} and Figure \ref{fig:classification}) and training process described in Section \ref{i3d+lstm}.

\section{Experiments and Results} \label{experiments}
\subsection{Evaluation method}
To assess the effectiveness of our proposed methods for equine ear movement detection, we conducted a series of experiments evaluating movement presence classification across dataset clips. The dataset consists of 283 clips of varying lengths (0.5s-3s), with 135 clips containing ear movements and 145 representing background (no ear movement). Expert annotated EquiFACS ear related AUs or EADs labels served as ground truth for classification (from \cite{rashidEquineFacialAction2020}).

Each method was evaluated using a test set after parameter optimization on training data. Hyperparameters, such as feature extraction window size, step size, frames per second (FPS), and LSTM architecture details, such as number of layers (\# layers), hidden size, and learning rate (lr) were systematically fine-tuned to maximize accuracy on train/validation dataset. Following configurations from Tables \ref{tab:feature-extraction} and \ref{tab:train-config}, we trained multiple models with different hyperparameters and kept the ones that performed best on validation for testing (see Table \ref{tab:results}).

\begin{table*}
\small
\centering
\begin{tabular}{lccccc}
\toprule
\multicolumn{1}{l}{\textbf{Method}} &
  \multicolumn{1}{c}{\textbf{Streams}} &
  \textbf{FPS} &
  \textbf{Sample rate} &
  \multicolumn{1}{c}{\textbf{Window}} &
  \textbf{Step} \\ \midrule
\multicolumn{1}{l}{\textbf{I3D+LSTM}} &
  \multicolumn{1}{c}{RGB, Flow, Mixed} &
  {[}25, 50{]} &
  - &
  \multicolumn{1}{c}{{[}32, 16{]}} &
  {[}16,8,1{]} \\ \midrule
\multicolumn{1}{l}{\textbf{VideoMAE+LSTM}} &
  \multicolumn{1}{c}{RGB} &
  {[}25, 50{]} &
  {[}1,2,4,8{]} &
  \multicolumn{1}{c}{-} &
  - \\ \bottomrule
\end{tabular}
\caption{Feature extraction experimental configuration setup. All configurations were used in training via grid-search.}
\label{tab:feature-extraction}
\end{table*}

\begin{table*}
\small
\centering
\begin{tabular}{lcccc}
\toprule
\textbf{Method}        & \textbf{Feature size} & \textbf{\# Layers} & \textbf{Hidden size} & \textbf{Learning rate (lr)}       \\ \midrule
\textbf{I3D+LSTM}      & 1024                  & [2,3]             & [256, 512]           & [0.0005, 0.001, 0.005, 0.01] \\ \midrule
\textbf{VideoMAE+LSTM} & 768                   & [2,3]             & [256, 512]           & [0.0005, 0.001, 0.005, 0.01] \\ \bottomrule
\end{tabular}
\centering
\caption{Experimental LSTM training configurations. All configurations were used in training via grid search. Best configurations were selected for testing (see Table \ref{tab:results})}
\label{tab:train-config}
\end{table*}

\subsection{Quantitative results}

Table \ref{tab:results} presents the classification accuracy and F1-score for each method’s best-performing configuration. We selected the models that achieved the best validation accuracy for each method.

Our optical flow-based approach (movDet) achieved an accuracy of 0.75 and an F1-score of 0.739, indicating moderate effectiveness in detecting ear movements from the video data. The I3D+LSTM model, leveraging deep spatiotemporal features, significantly improved upon this baseline, reaching 0.8125 accuracy and an F1-score of 0.816. Finally, the VideoMAE+LSTM model outperformed both, attaining the highest accuracy of 0.875 and an F1-score of 0.869, demonstrating the efficacy of VideoMAE's transformer-based spatiotemporal representations in this context.

\begin{table*}
\centering
\small
\begin{tabular}{lccccccccc}
\toprule
\multicolumn{1}{l}{\textbf{Method}} &
  \textbf{FPS} &
  \multicolumn{1}{c}{\textbf{\# Layers}} &
  \textbf{Hidden size} &
  \textbf{lr} &
  \textbf{Sample rate} &
  \multicolumn{1}{c}{\textbf{Window}} &
  \textbf{Step} &
  \multicolumn{1}{c}{\textbf{Accuracy}} &
  \multicolumn{1}{c}{\textbf{F1}} \\ \midrule
\multicolumn{1}{l}{\textbf{movDet (Flow)}} &
  25 &
  \multicolumn{1}{c}{-} &
  - &
  - &
  - &
  \multicolumn{1}{c}{-} &
  - &
  \multicolumn{1}{c}{\textbf{0.75}} &
  \multicolumn{1}{c}{\textbf{0.73913}} \\ \midrule
\multicolumn{1}{l}{\textbf{I3D+LSTM (Flow)}} &
  50 &
  \multicolumn{1}{c}{3} &
  256 &
  0.001 &
  - &
  \multicolumn{1}{c}{32} &
  16 &
  \multicolumn{1}{c}{\textbf{0.8125}} &
  \multicolumn{1}{c}{\textbf{0.81633}} \\ \midrule
\multicolumn{1}{l}{\textbf{I3D+LSTM (Mixed)}} &
  50 &
  \multicolumn{1}{c}{3} &
  256 &
  0.005 &
  - &
  \multicolumn{1}{c}{32} &
  16 &
  \multicolumn{1}{c}{0.75} &
  \multicolumn{1}{c}{0.76923} \\ \midrule
\multicolumn{1}{l}{\textbf{I3D+LSTM (RGB)}} &
  50 &
  \multicolumn{1}{c}{3} &
  256 &
  0.005 &
  - &
  \multicolumn{1}{c}{32} &
  16 &
  \multicolumn{1}{c}{0.625} &
  \multicolumn{1}{c}{0.67857} \\ \midrule
\multicolumn{1}{l}{\textbf{VideoMAE+LSTM (RGB)}} &
  50 &
  \multicolumn{1}{c}{2} &
  256 &
  0.001 &
  8 &
  \multicolumn{1}{c}{-} &
  - &
  \multicolumn{1}{c}{\textbf{0.875}} &
  \multicolumn{1}{c}{\textbf{0.86957}} \\ \bottomrule
\end{tabular}
\caption{Test set results on movDet, I3D+LSTM and VideoMAE+LSTM. Best stream configuration accuracy results are presented for each method, when applicable.}
\label{tab:results}
\end{table*}

\begin{figure}
    \centering
    \begin{subfigure}{0.2\textwidth}
        \centering
        \includegraphics[width=\textwidth]{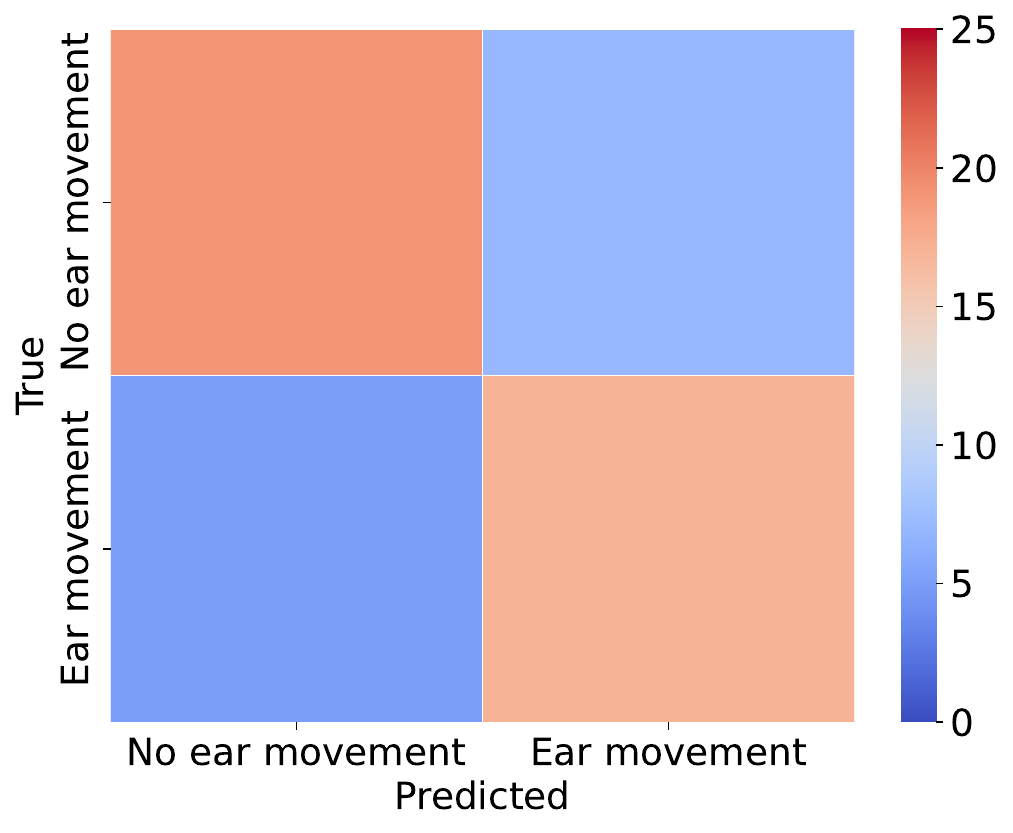}
        \caption{movDet}
    \end{subfigure}
    \begin{subfigure}{0.2\textwidth}
        \centering
        \includegraphics[width=\textwidth]{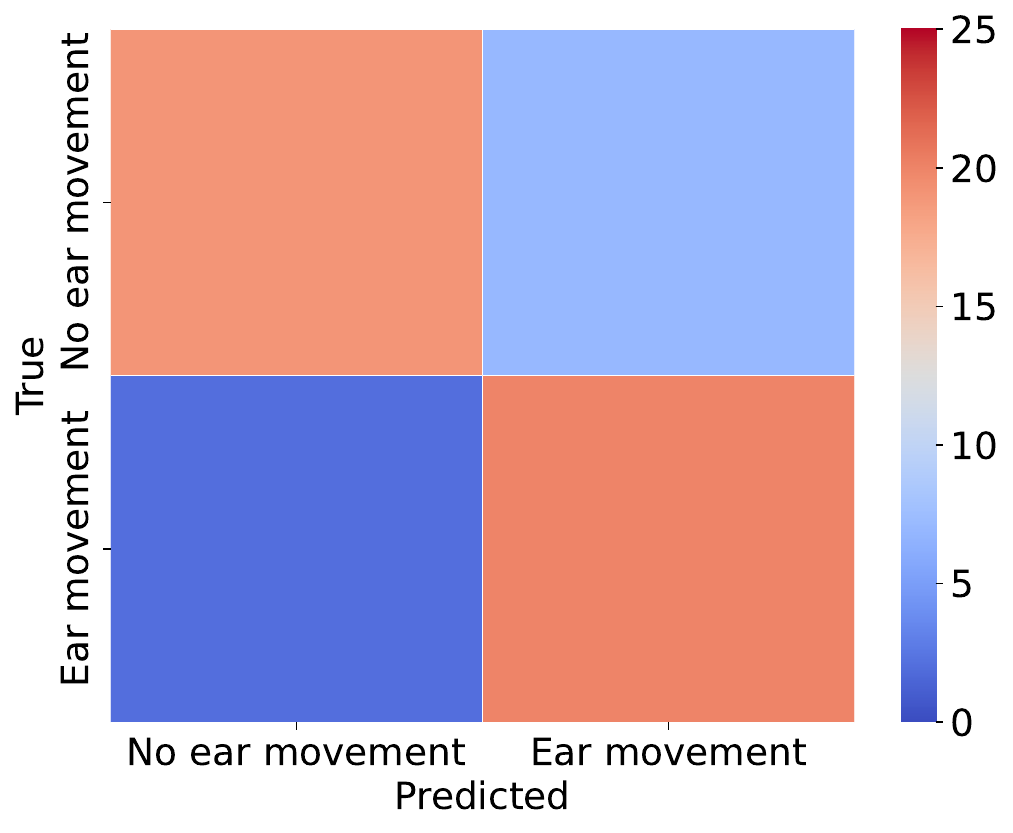}
        \caption{I3D+LSTM}
    \end{subfigure}
    \begin{subfigure}{0.2\textwidth}
        \centering
        \includegraphics[width=\textwidth]{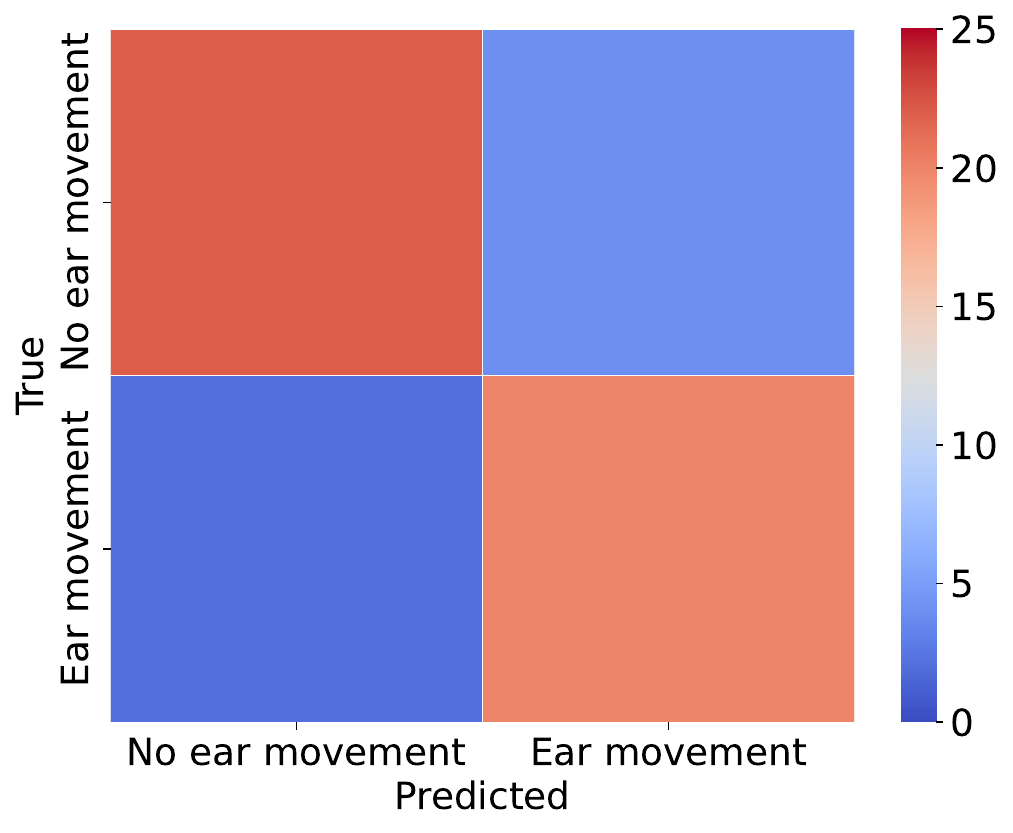}
        \caption{VideoMAE+LSTM}
    \end{subfigure}
    \caption{Confusion matrices for test set evaluation for each method.}
    \label{fig:CM}
\end{figure}

Figure \ref{fig:CM} shows confusion matrices for each method, illustrating the classification performance. The movDet approach suffered from a higher false positive rate, whereas both deep learning-based methods exhibited greater precision and recall. Notably, the VideoMAE+LSTM model achieved highest accuracy.

\subsection{Qualitative results}
\begin{figure}
    \centering
    \begin{subfigure}[b]{0.8\linewidth} 
        \centering
        \includegraphics[width=\textwidth]{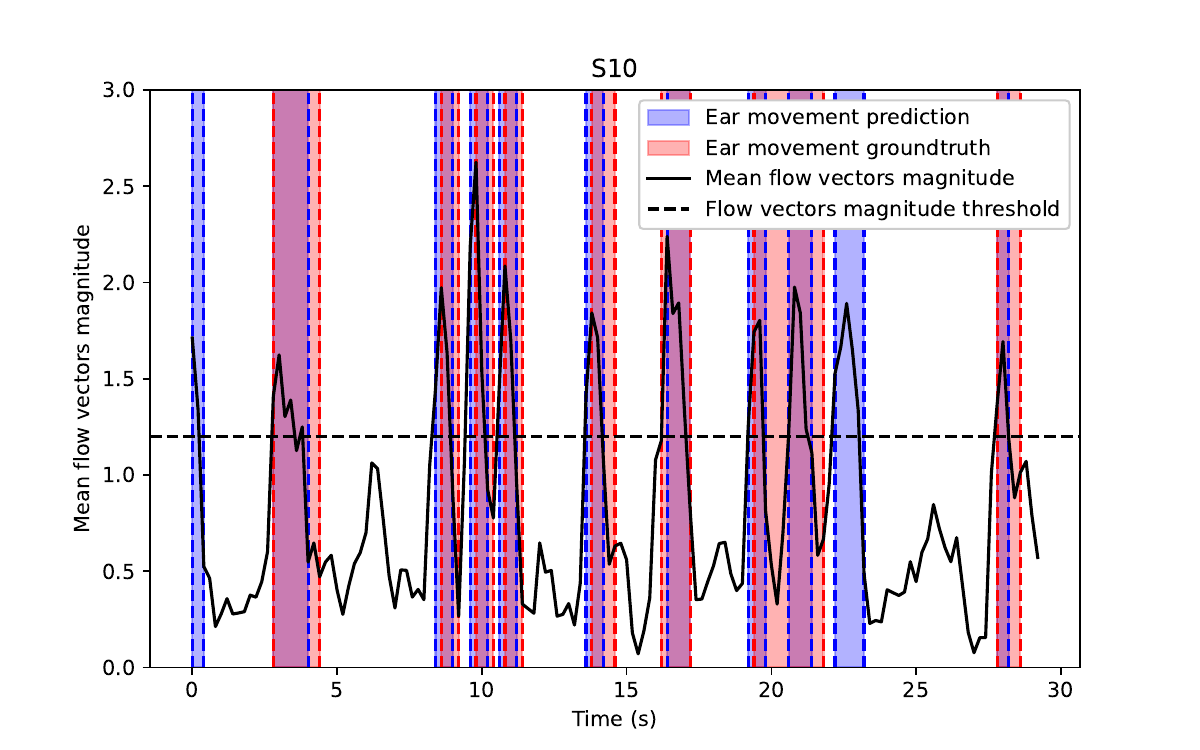}
        \caption{Ear movement instances detection on video S10 \cite{rashidEquineFacialAction2020}.}
        \label{fig:movdetsuccess}
    \end{subfigure}
    \begin{subfigure}[b]{0.8\linewidth}
        \centering
        \includegraphics[width=\textwidth]{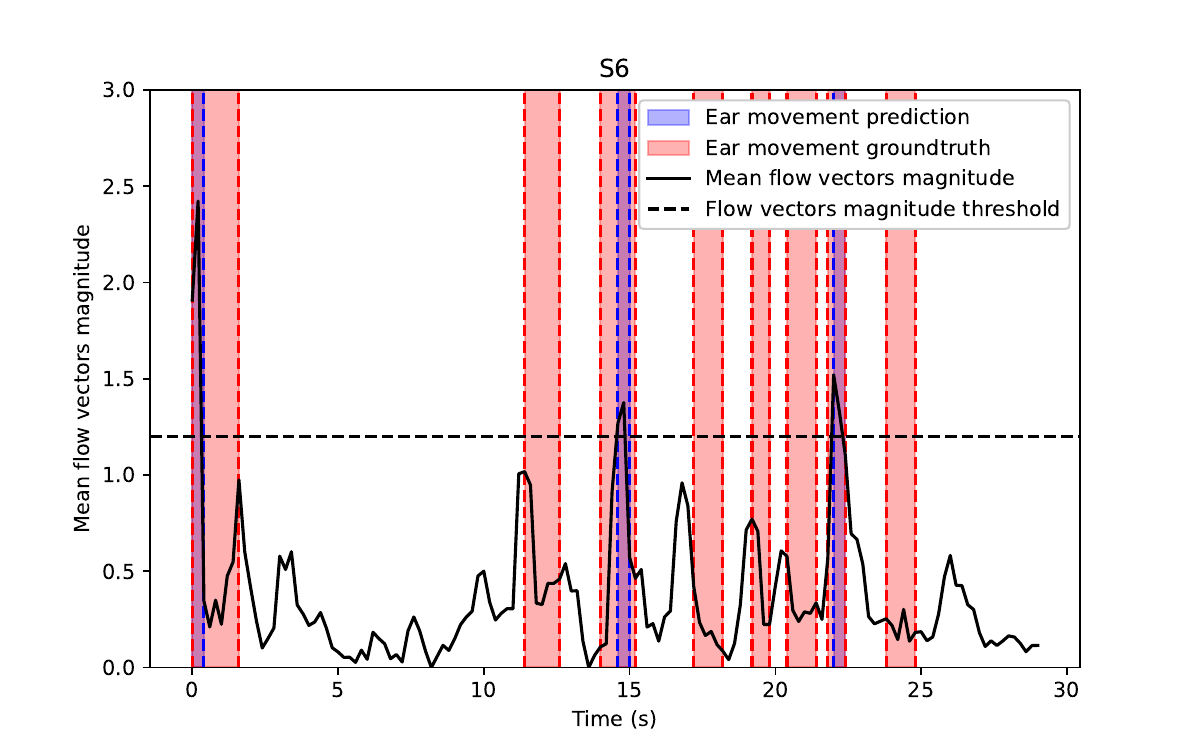}
        \caption{Ear movement instance detection failure on video S6 \cite{rashidEquineFacialAction2020}.}
        \label{fig:movdetfail}
    \end{subfigure}
    \caption{Qualitative analysis for movDet method on original full-length horse videos. Additional results can be found
in the provided supplementary material.}
    \label{fig:qual_movdet}
\end{figure}

\begin{figure}
    \centering
    \begin{subfigure}[b]{0.8\linewidth} 
        \centering
        \includegraphics[width=\textwidth]{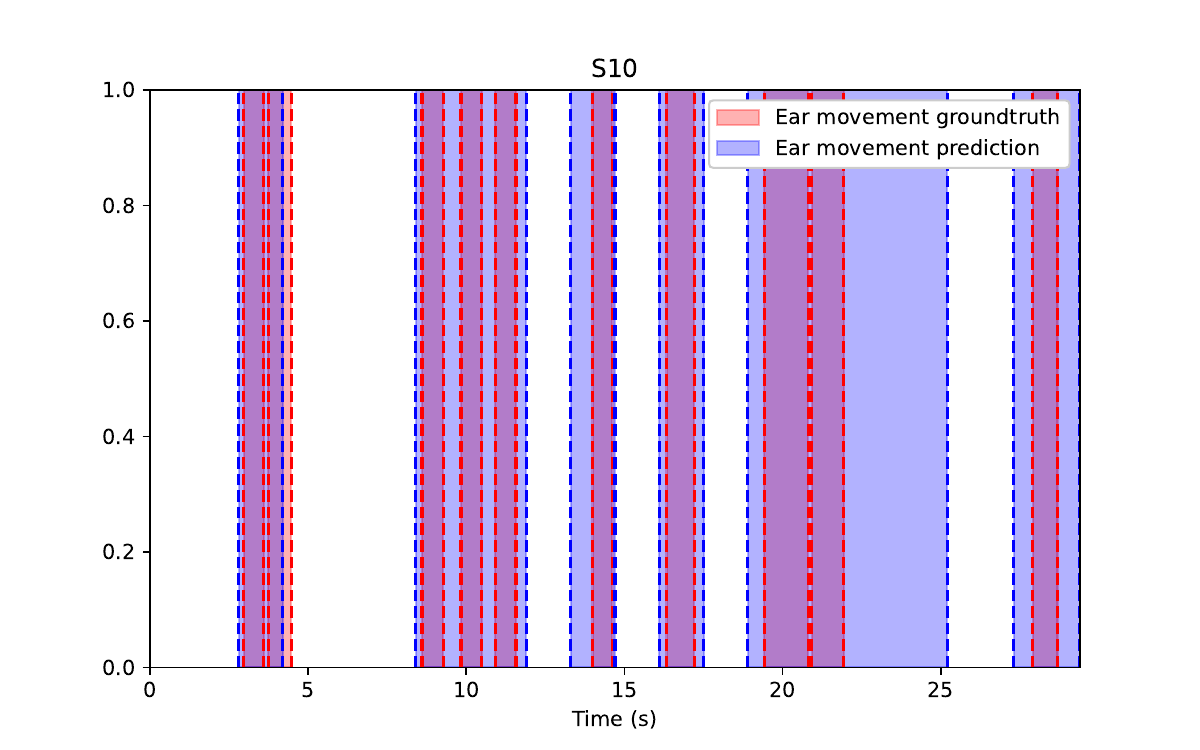}
        \caption{Ear movement instances detection on video S10 \cite{rashidEquineFacialAction2020}.}
    \end{subfigure}
    \begin{subfigure}[b]{0.8\linewidth}
        \centering
        \includegraphics[width=\textwidth]{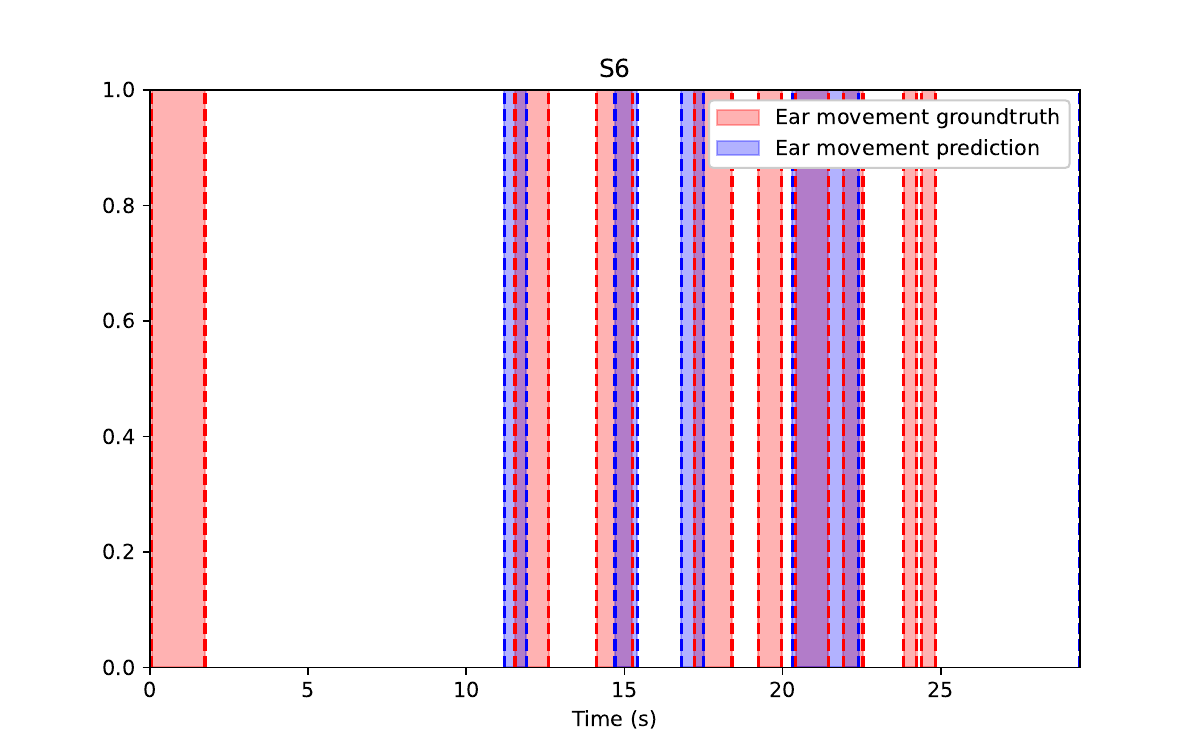}
        \caption{Ear movement instance detection on video S6 \cite{rashidEquineFacialAction2020}.}
    \end{subfigure}
    \caption{Qualitative analysis for I3D+LSTM method on original full-length horse videos. Additional results can be found in the provided supplementary material.}
    \label{fig:qual_I3D}
\end{figure}

\begin{figure}
    \centering
    \begin{subfigure}[b]{0.8\linewidth} 
        \centering
        \includegraphics[width=\textwidth]{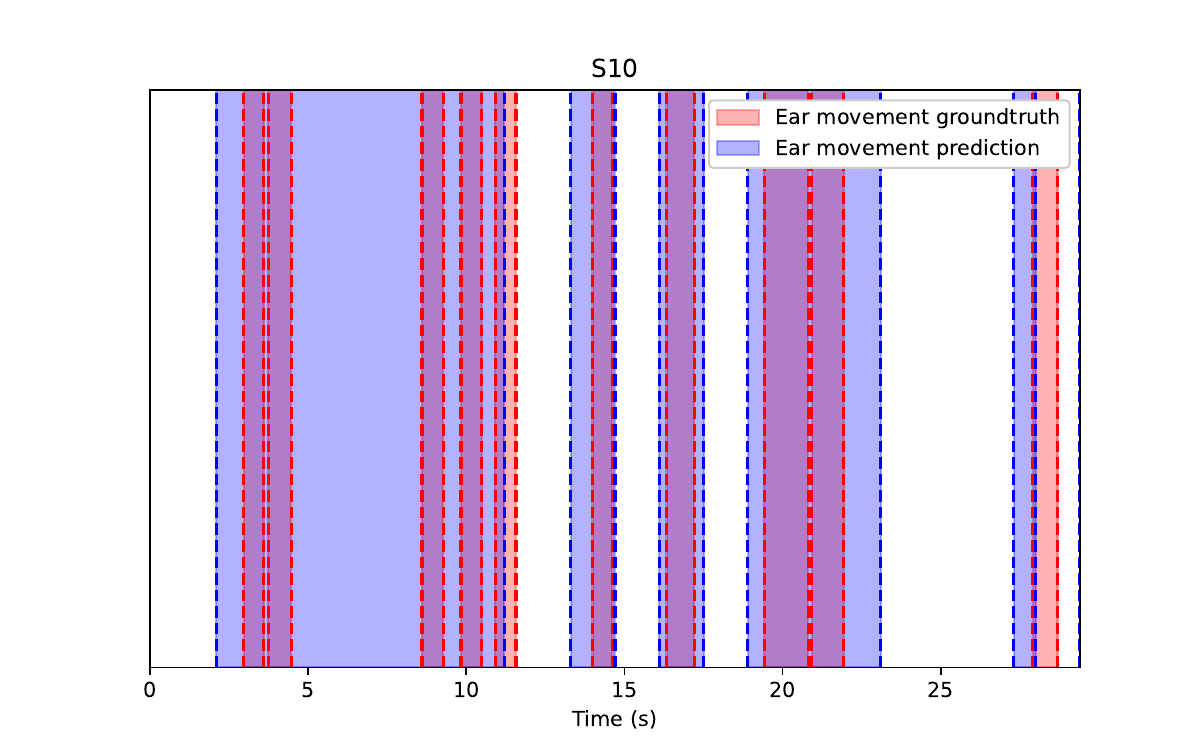}
        \caption{Ear movement instances detection on video S10 \cite{rashidEquineFacialAction2020}.}
    \end{subfigure}
    \begin{subfigure}[b]{0.8\linewidth}
        \centering
        \includegraphics[width=\textwidth]{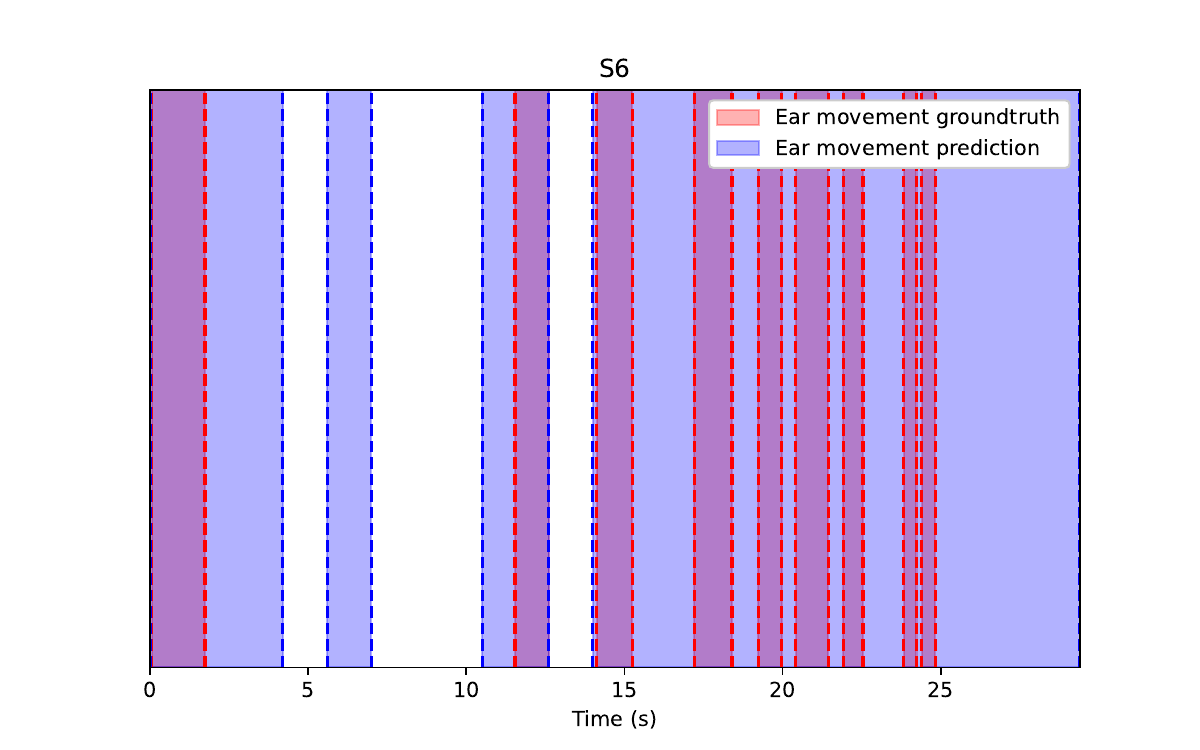}
        \caption{Ear movement instance detection on video S6 \cite{rashidEquineFacialAction2020}.}
    \end{subfigure}
    \caption{Qualitative analysis for VideoMAE+LSTM method on original full-length horse videos. Additional results can be found in the provided supplementary material.}
    \label{fig:qual_videomae}
\end{figure}

To further evaluate performance in a real world application, we naively classify ear movement with a window-based approach in the original full-length horse videos using movDet, I3D+LSTM and VideoMAE+LSTM. For the latter two models, the videos are segmented into overlapping clips of 50 frames, with a stride of 35 frames between windows. Qualitative results can be found for two of the full-length videos in Figures \ref{fig:qual_movdet}, \ref{fig:qual_I3D} and \ref{fig:qual_videomae}. 

\section{Discussion}
Our study demonstrates the potential of deep learning-driven automation for equine pain assessment, particularly in Action Unit (AU) detection, despite the scarcity of publicly available data. Compared to our optical flow-based baseline, both the I3D+LSTM and VideoMAE+LSTM models significantly improve accuracy by leveraging spatiotemporal deep learning features. Notably, our results with the transformer-based VideoMAE feature extractor combined with a simple RNN classifier (LSTM) suggest that transformer architectures may be instrumental in addressing the limited availability of annotated datasets, which is one of the primary challenges in equine affective computing. While challenges remain, we believe this work represents a meaningful step toward more robust and scalable AU detection.
\section{Conclusion}
These findings lay the groundwork for further advancements in automated AU localization for equine welfare monitoring and highlight a promising pipeline for cross-species applications. The methodologies developed here could be adapted for other species with similar affective state indicators, advancing animal welfare monitoring across various domains. Future research should continue exploring transformer-based models to enhance real-world applicability and improve the accuracy and efficiency of automated action unit detection systems, ultimately fostering a broader understanding of affective states across different species. The code and data used in this work will be made publicly available upon paper publication.
\paragraph{Acknowledgements.} This work has been funded by the Independent Research Fund Denmark under grant ID 10.46540/3105-00114B.

\clearpage
{
    \small
    \bibliographystyle{ieeenat_fullname}
    \bibliography{bibliography}
}
\clearpage
\setcounter{page}{1}
\maketitlesupplementary

This document contains the supplementary material for CVPR 2025 ABAW Workshop Paper \#51 and provides further insight into the results obtained with the three different methods tested (movDet, I3D+LSTM and VideoMAE+LSTM).

\section{Dataset subjects sample data}
\begin{figure*}[htbp]
    \centering

    \begin{subfigure}{0.32\linewidth}
        \includegraphics[width=\linewidth]{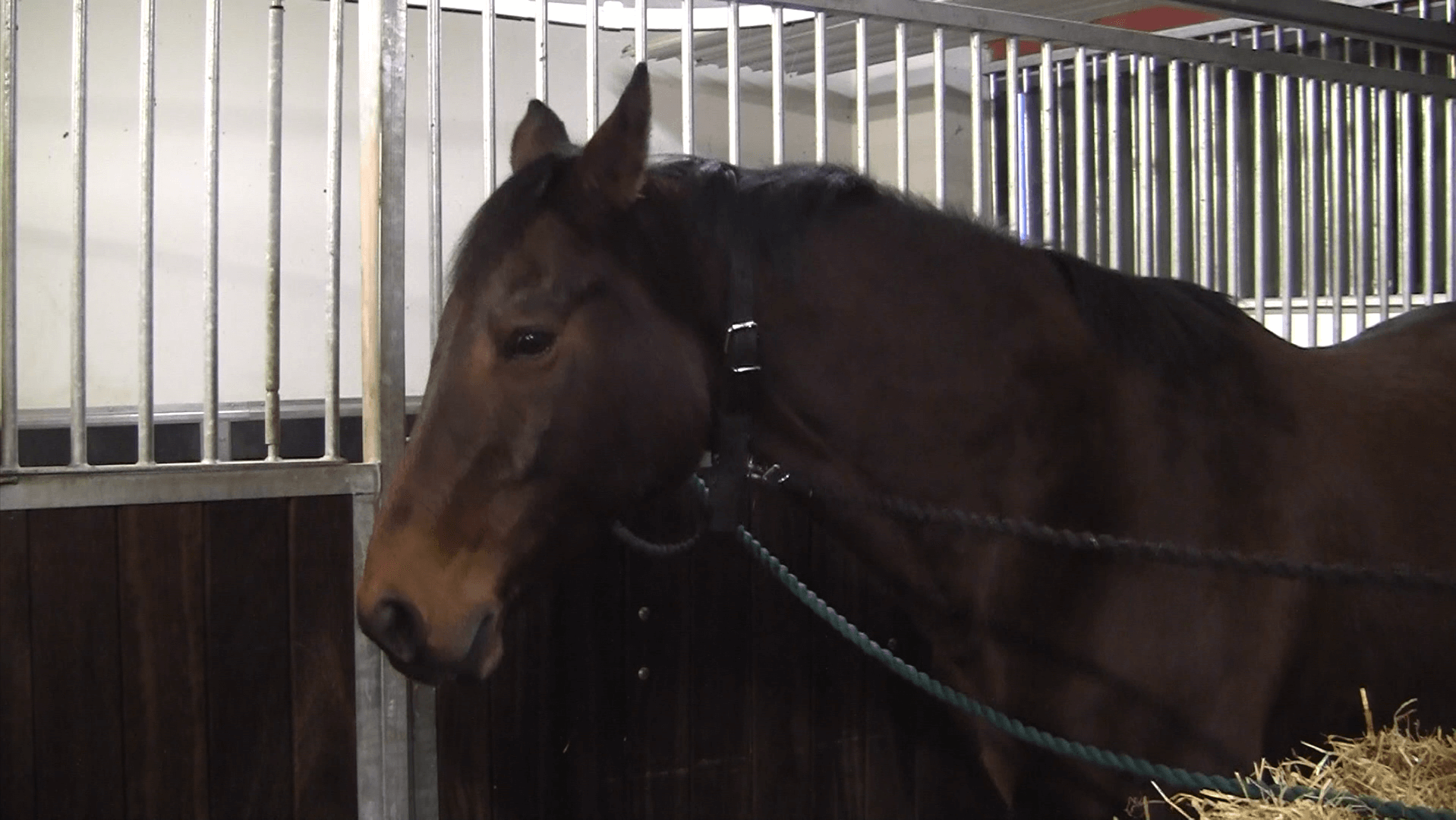}
    \end{subfigure}
    \begin{subfigure}{0.32\linewidth}
        \includegraphics[width=\linewidth]{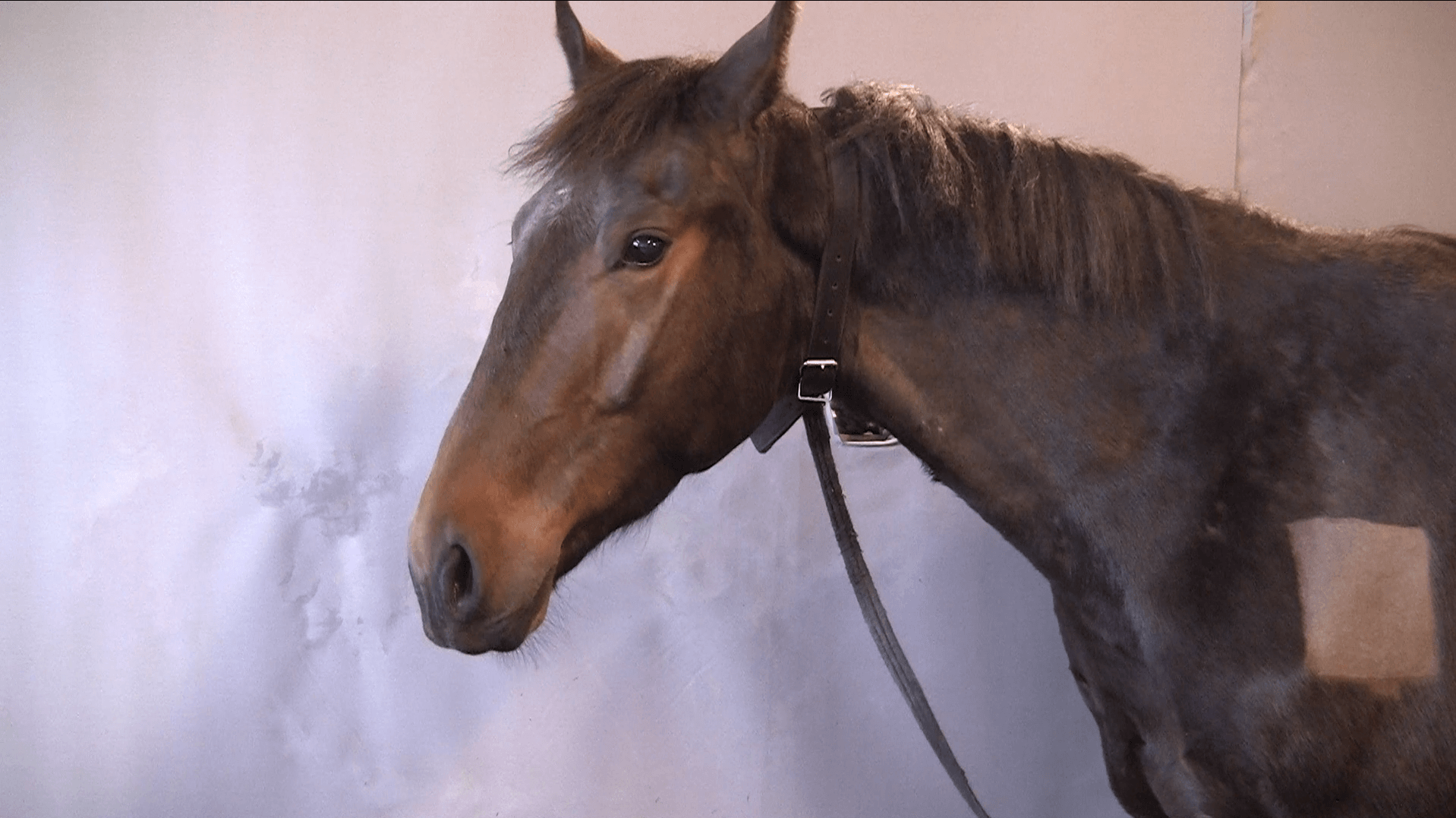}
    \end{subfigure}
    \begin{subfigure}{0.32\linewidth}
        \includegraphics[width=\linewidth]{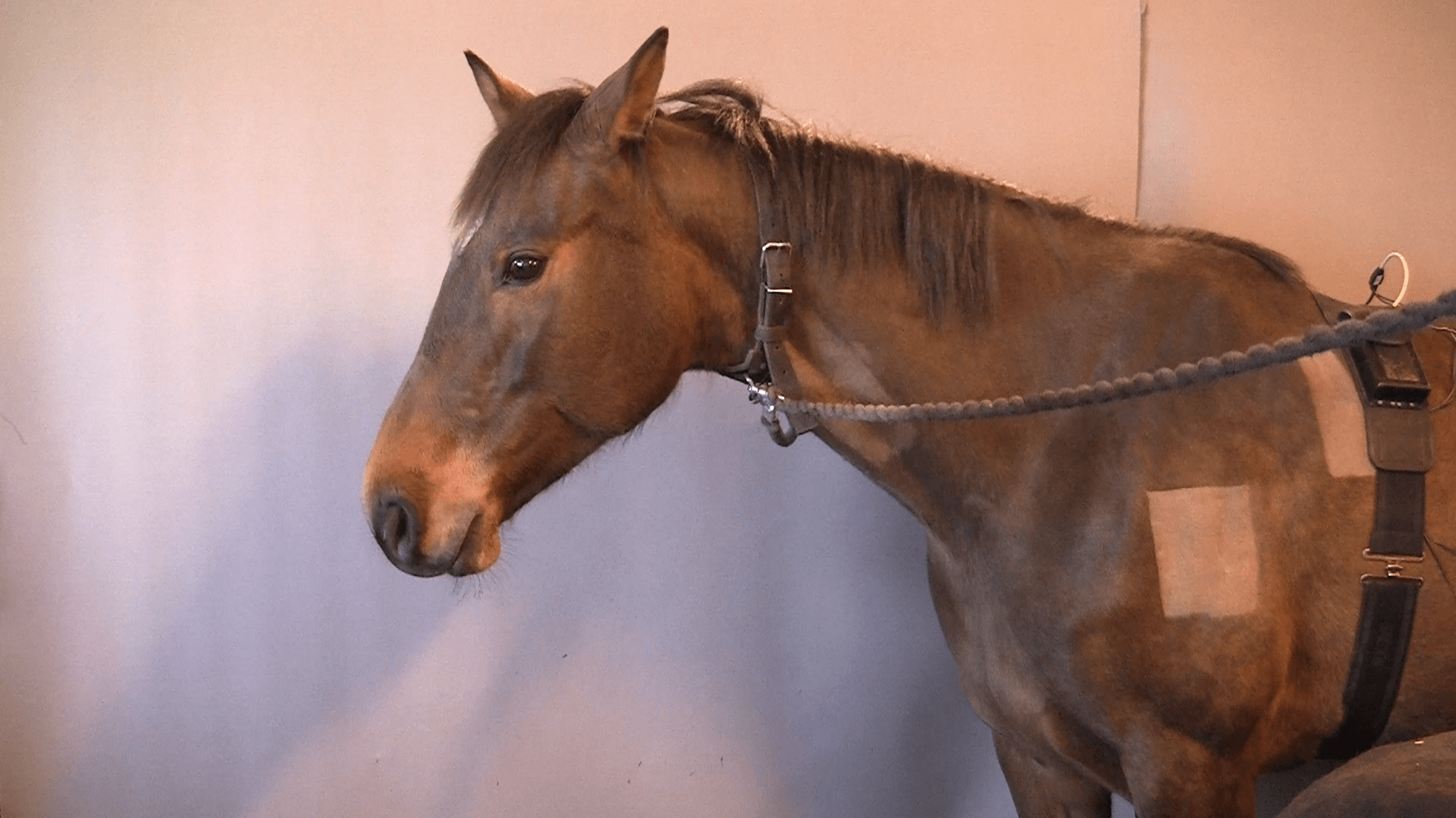}
    \end{subfigure}
    \\[0.5em]

    \begin{subfigure}{0.32\linewidth}
        \includegraphics[width=\linewidth]{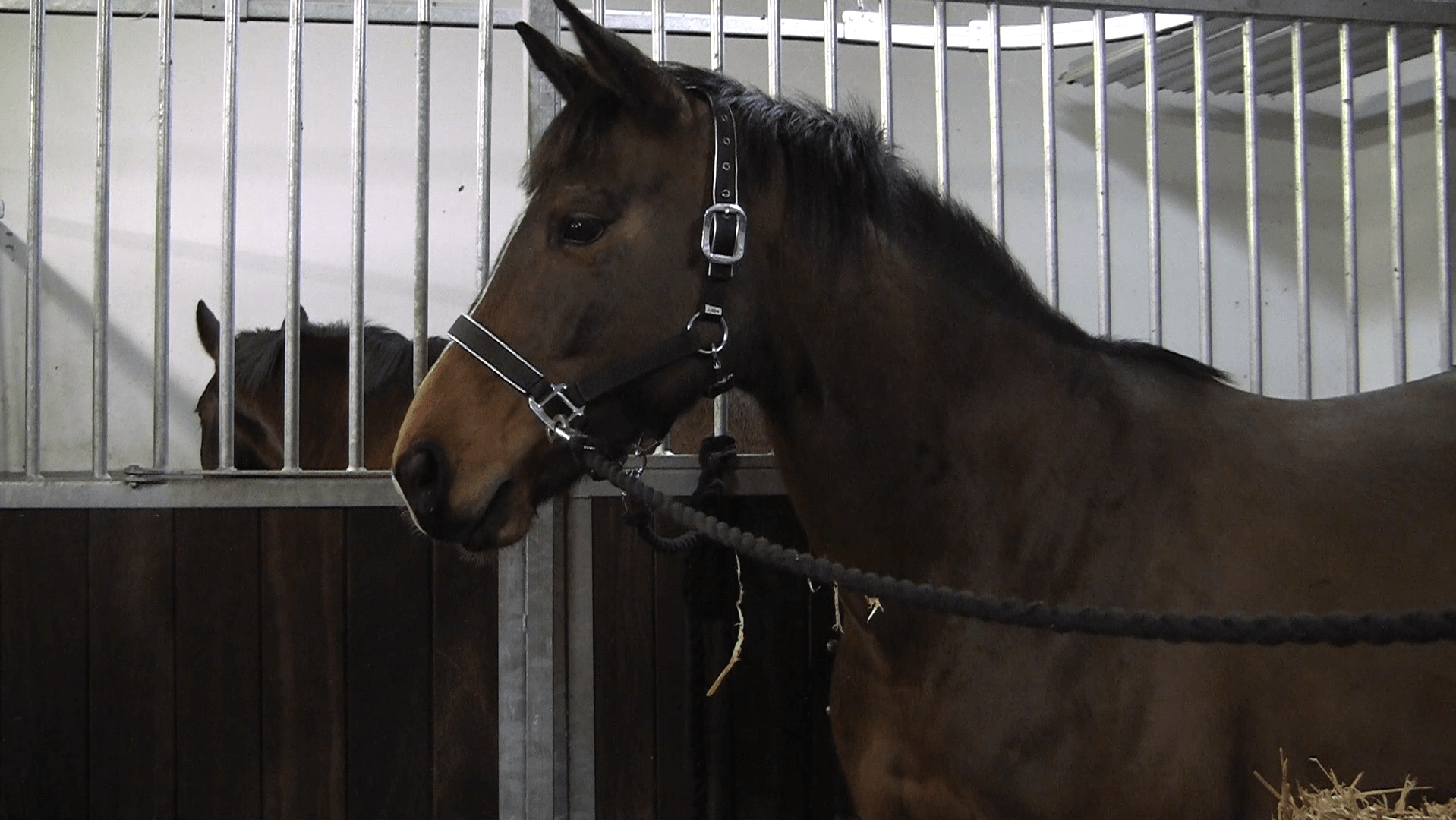}
    \end{subfigure}
    \begin{subfigure}{0.32\linewidth}
        \includegraphics[width=\linewidth]{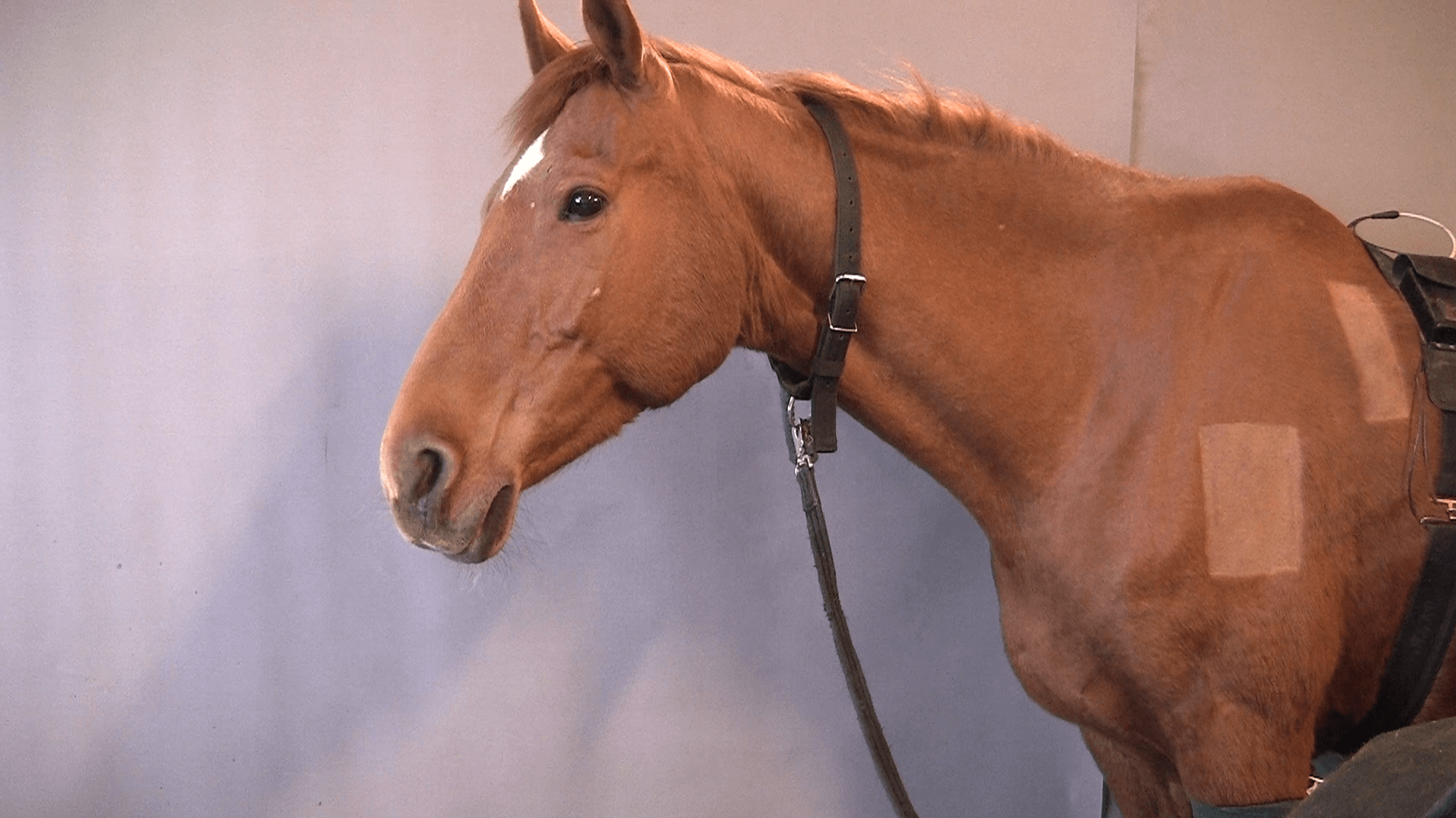}
    \end{subfigure}
    \begin{subfigure}{0.32\linewidth}
        \includegraphics[width=\linewidth]{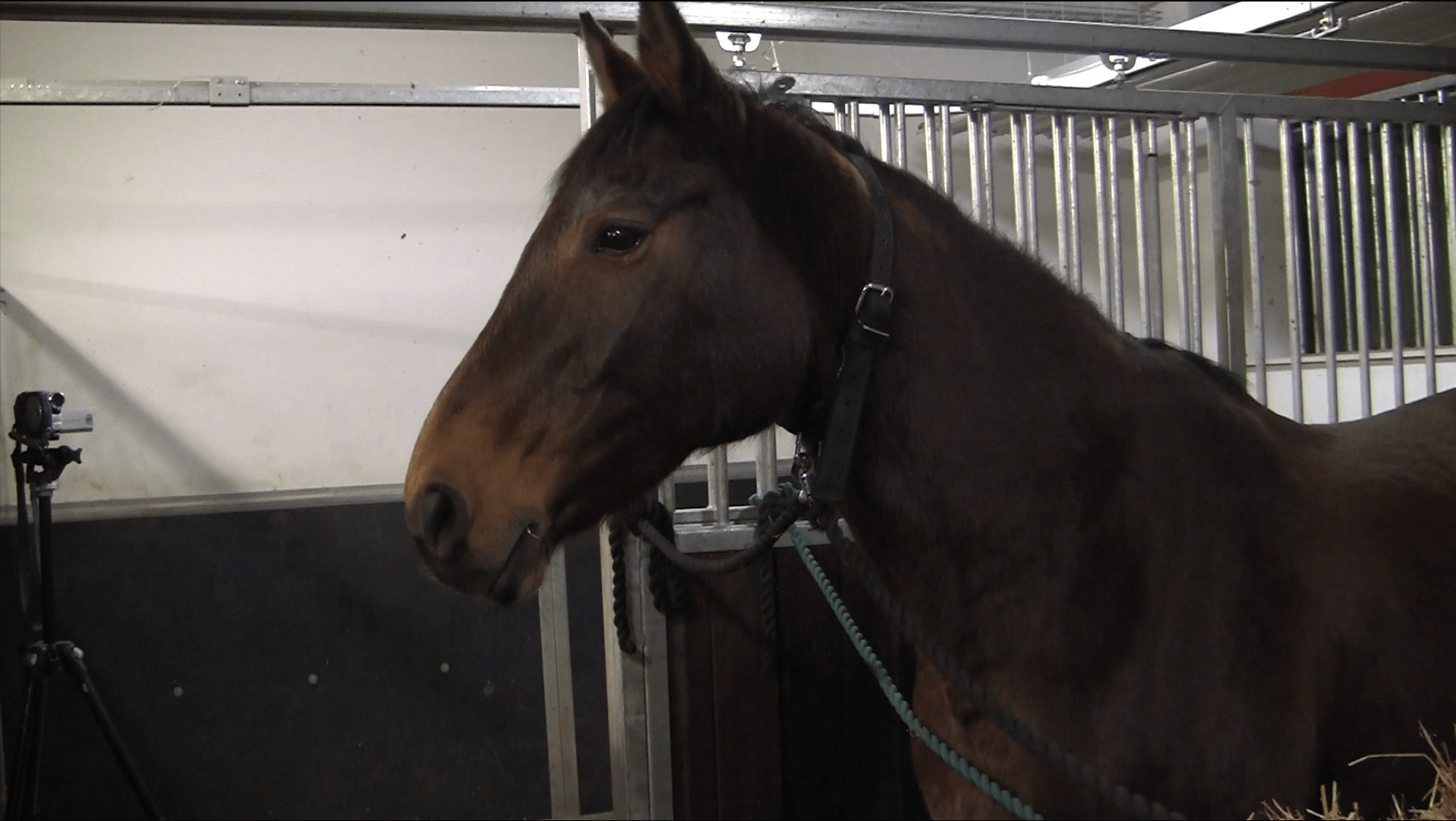}
    \end{subfigure}
    \\[0.5em]

    \begin{subfigure}{0.32\linewidth}
        \includegraphics[width=\linewidth]{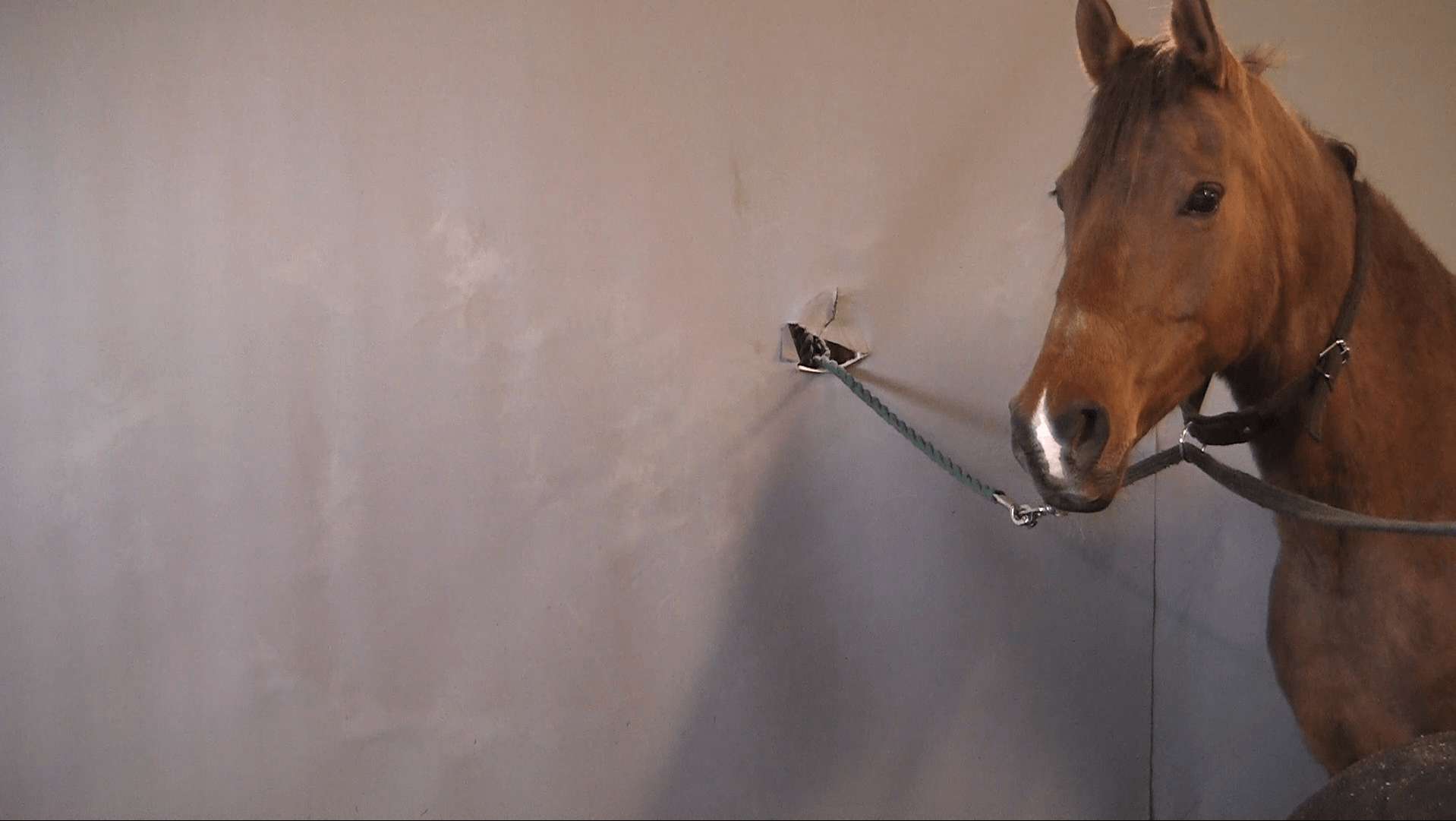}
    \end{subfigure}
    \begin{subfigure}{0.32\linewidth}
        \includegraphics[width=\linewidth]{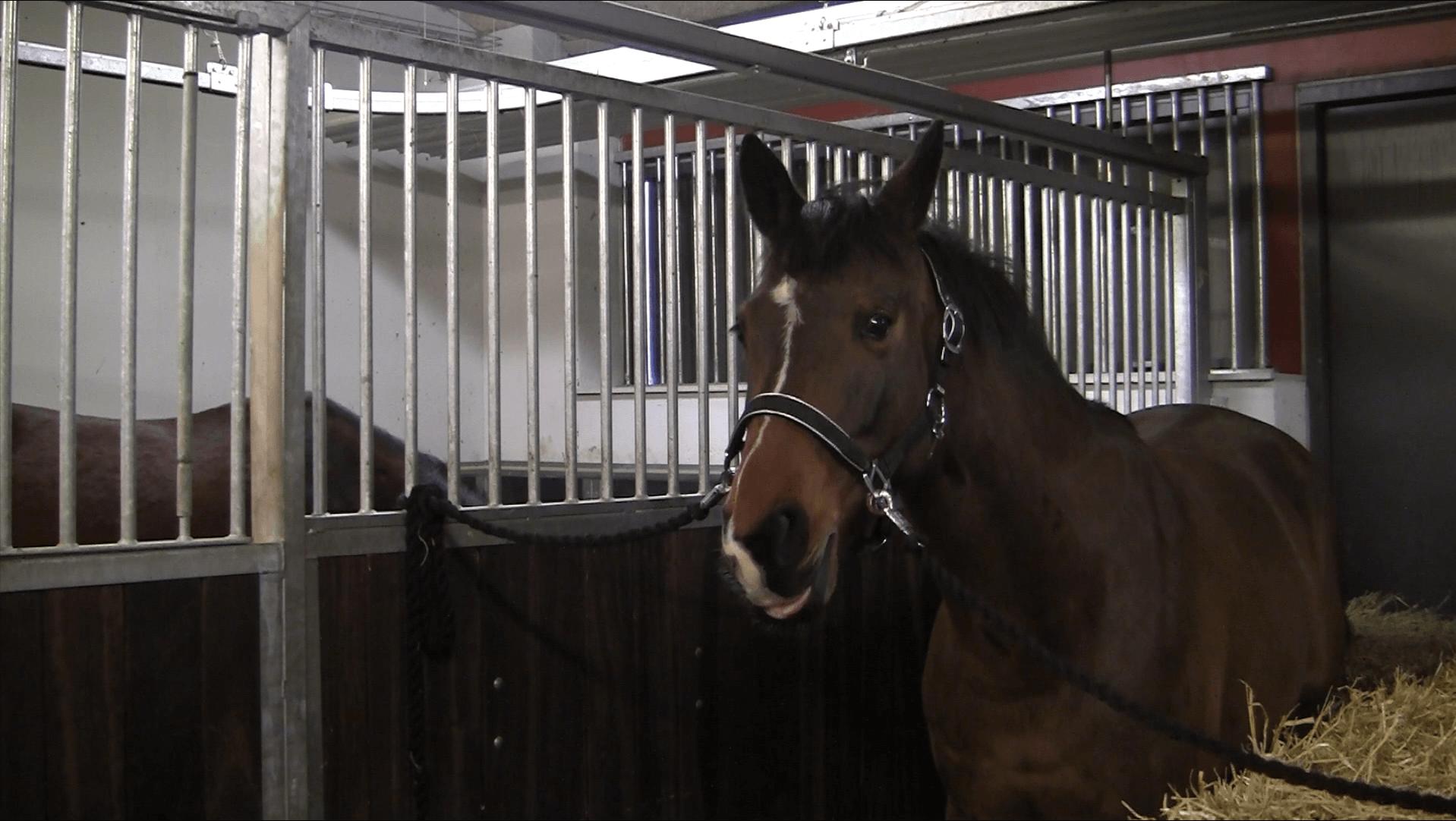}
    \end{subfigure}
    \begin{subfigure}{0.32\linewidth}
        \includegraphics[width=\linewidth]{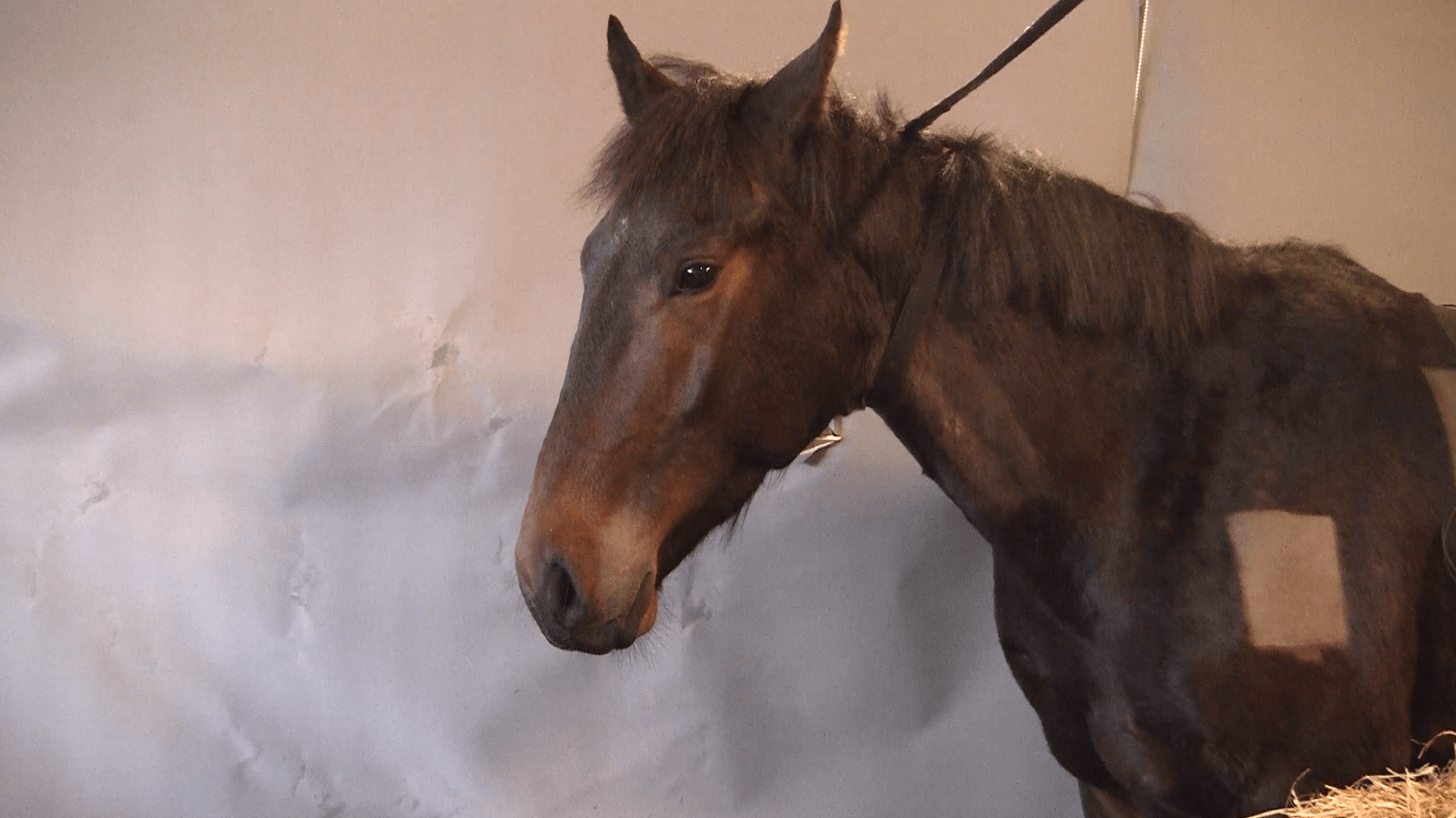}
    \end{subfigure}
    \\[0.5em]

    \begin{subfigure}{0.32\linewidth}
        \includegraphics[width=\linewidth]{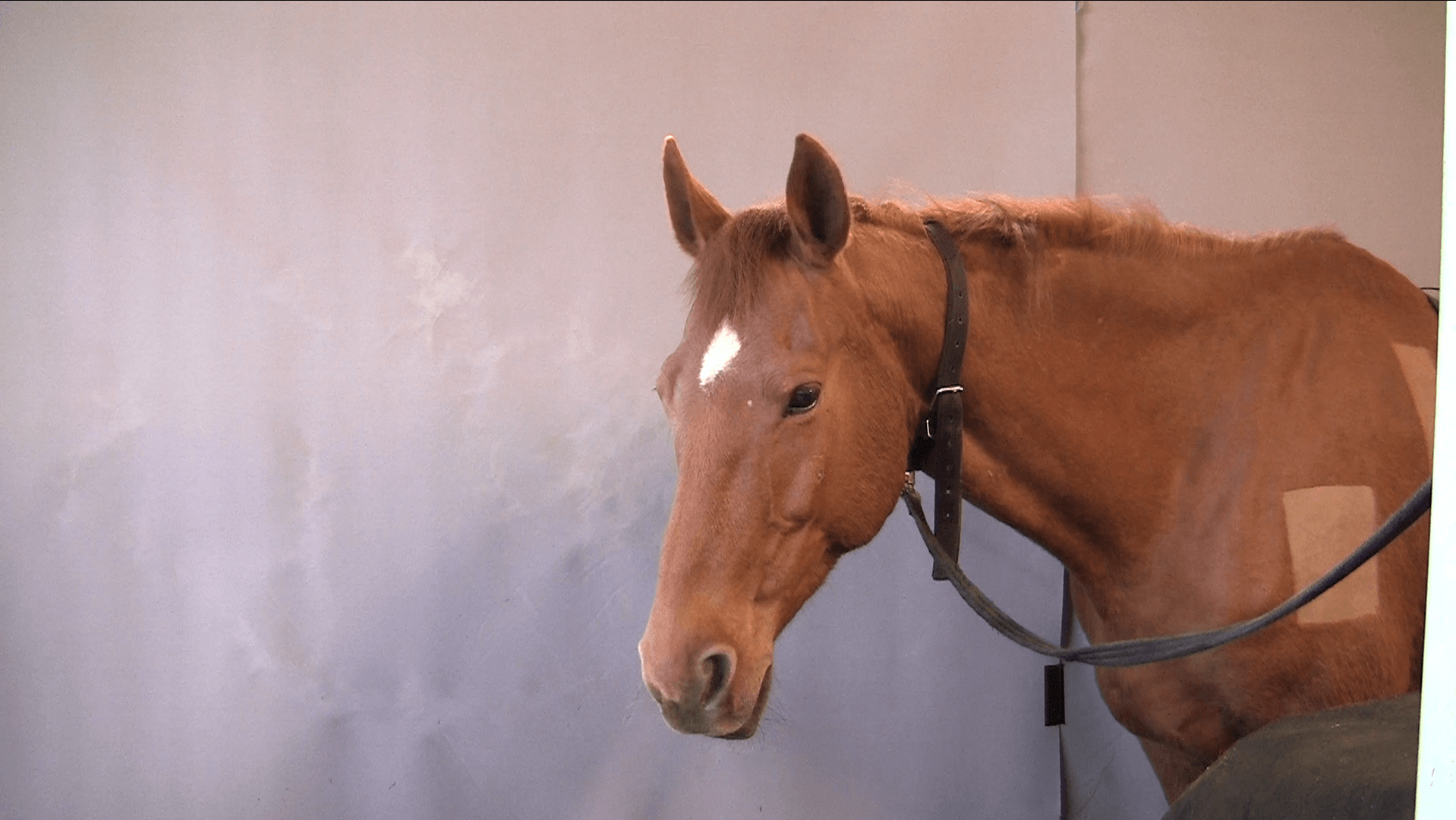}
    \end{subfigure}
    \begin{subfigure}{0.32\linewidth}
        \includegraphics[width=\linewidth]{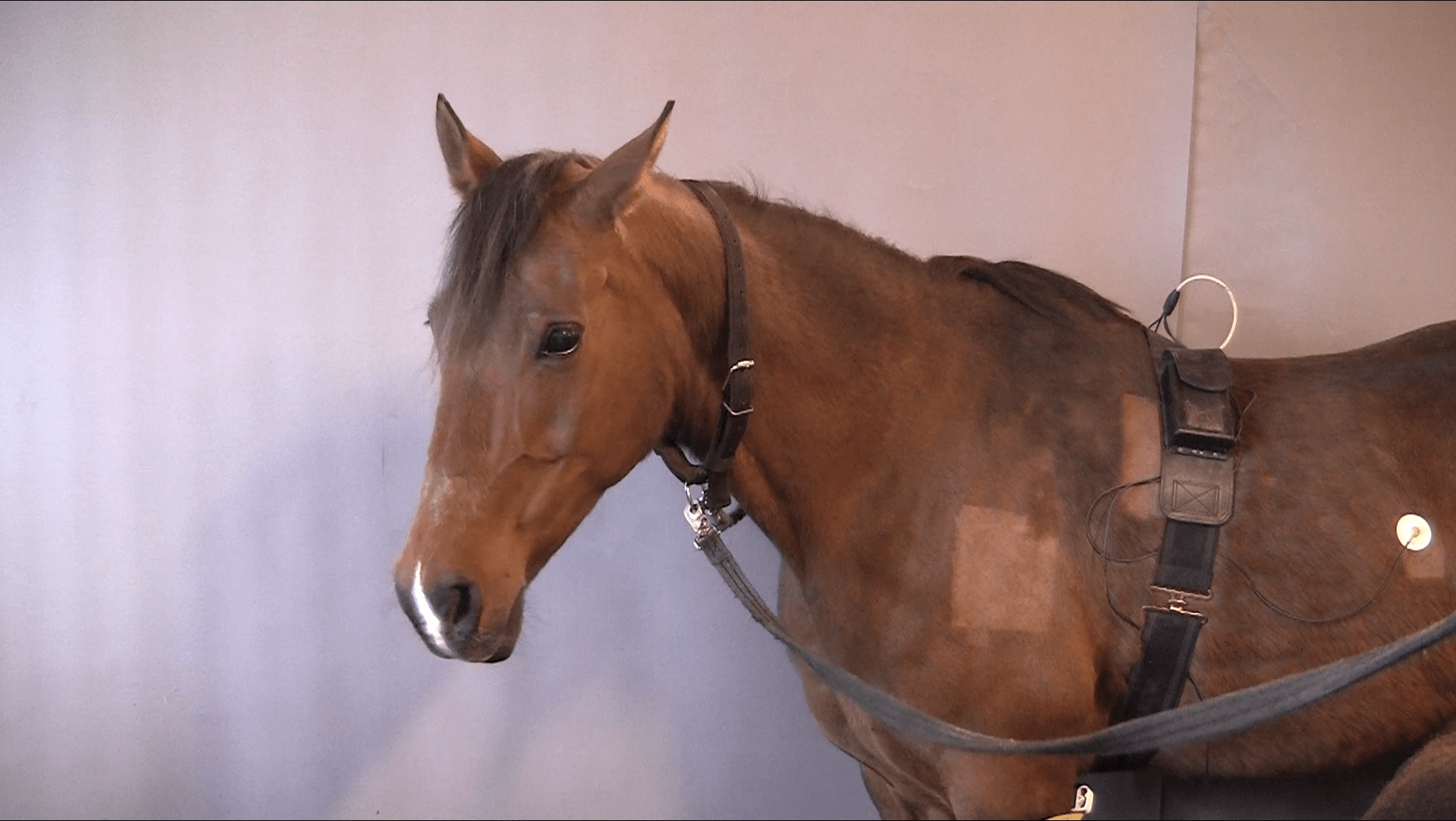}
    \end{subfigure}
    \begin{subfigure}{0.32\linewidth}
        \includegraphics[width=\linewidth]{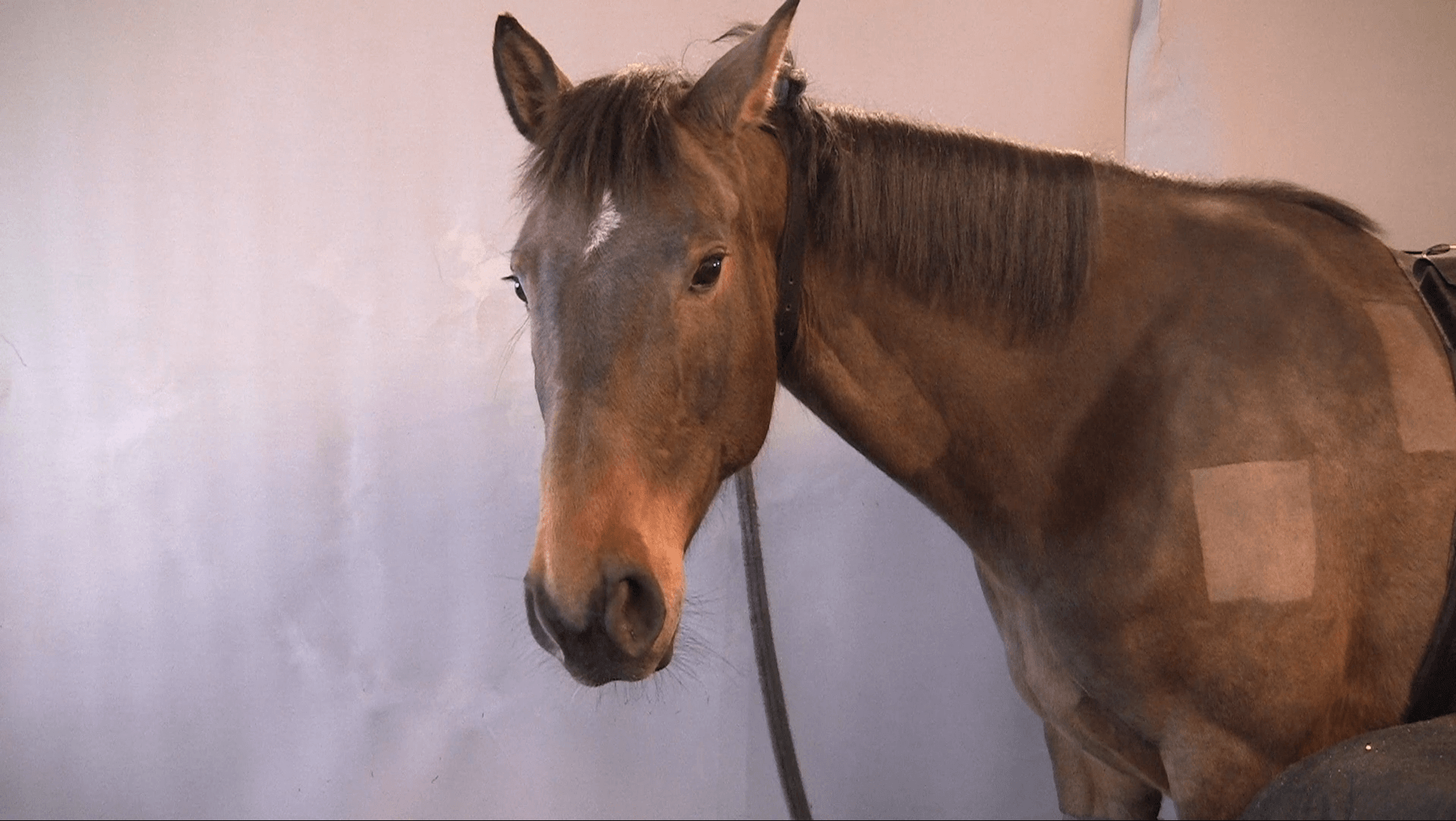}
    \end{subfigure}
    \caption{Sample frames for each of the 12 videos in the dataset in row-major order.}
    \label{fig:sample_frames}
\end{figure*}

To provide further insight into the dataset used, we provide sample  frames of each of the 12 videos in this supplementary material (see Figure \ref{fig:sample_frames}).

\section{Supplementary qualitative results}
In this section we show the qualitative results obtained from using each method on the original full length dataset videos.

\subsection{movDet}
MovDet works via sampling the video at a specific frame rate, then detecting and segmenting the horse's ear region. After that optical flow between both frames ear region is calculated. Finally we threshold the average magnitude of the flow vectors to obtain a ear movement/no-movement classification (see Figure \ref{fig:movDet}). We applies movDet directly to the original dataset RGB videos, obtaining time wise ear-movement classifications across each video. We condense these results into a single graph for each of the 12 videos in the dataset, where average flow gradient, groundtruth and predicted movement classification can be observed.

Figure \ref{fig:qual_supp_movdet} shows the qualitative results of the movDet method on the 12 dataset videos.

\subsection{I3D+LSTM}
For I3D+LSTM method, we adopted a window based approach to process the videos, selecting the top configuration tested from Table \ref{tab:results}. The method was applied to 50 FPS optical flow videos of the original data, using a window size of 50 frames and a stride of 35 frames. For each window we extracted the I3D flow stream features and classified it using the best configuration model. We condense these results into a single graph for each of the 12 videos in the dataset, where both groundtruth and predicted movement detection can be observed.

Figure \ref{fig:qual_supp_i3d_lstm} shows the qualitative results of the movDet method on the 12 dataset videos.

\subsection{VideoMAE+LSTM}
For VideoMAE+LSTM method, we adopted the same window based approach to process the videos, then selecting the top configuration tested from Table \ref{tab:results}. In this case, the method applied to 50 FPS RGB videos of the original data, using a window size of 50 frames and a stride of 35 frames. For each window we extracted the VideoMAE features and performed classification. As before, we condense these results into a single graph for each of the 12 videos in the dataset, where both groundtruth and predicted movement detection can be observed.

Figure \ref{fig:qual_supp_videomae_lstm} shows the qualitative results of the movDet method on the 12 dataset videos.

\clearpage 

\clearpage

\onecolumn

\section*{Qualitative Analysis: movDet}

\begin{figure}[H]
    \captionsetup{labelformat=empty}
    \centering
    \begin{subfigure}{0.49\linewidth}
        \includegraphics[width=\linewidth]{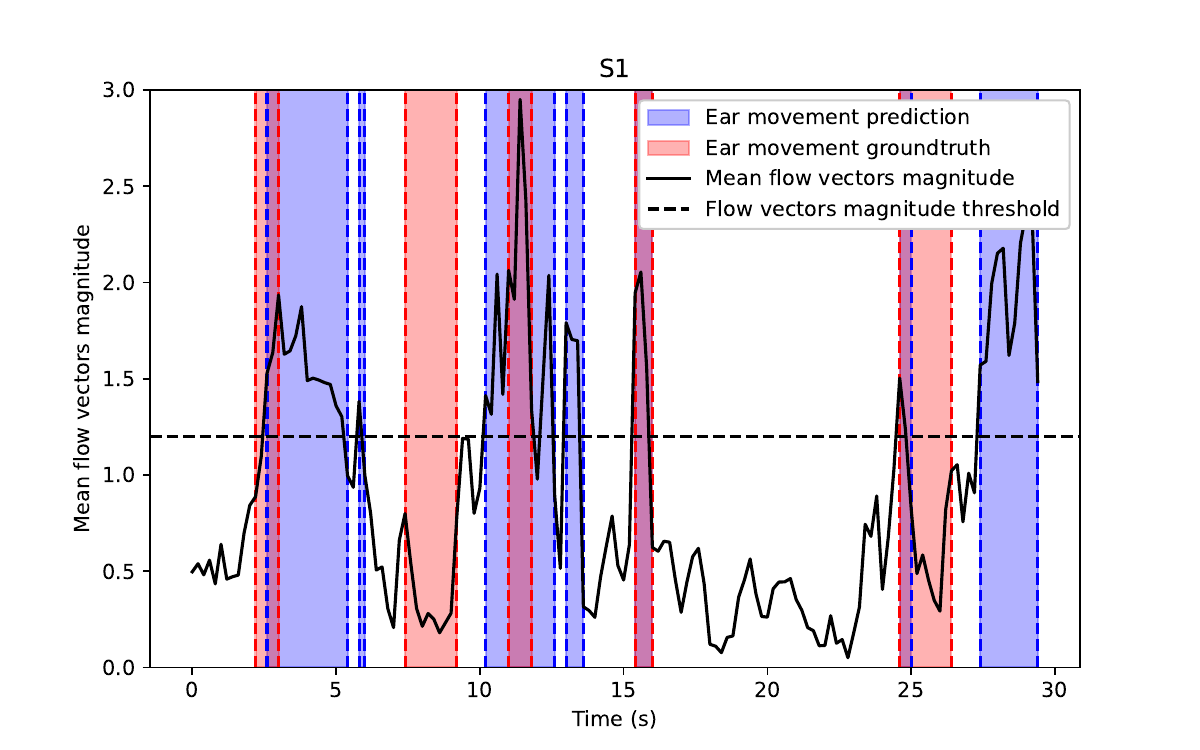}
        \caption{S1}
    \end{subfigure}
    \hfill
    \begin{subfigure}{0.49\linewidth}
        \includegraphics[width=\linewidth]{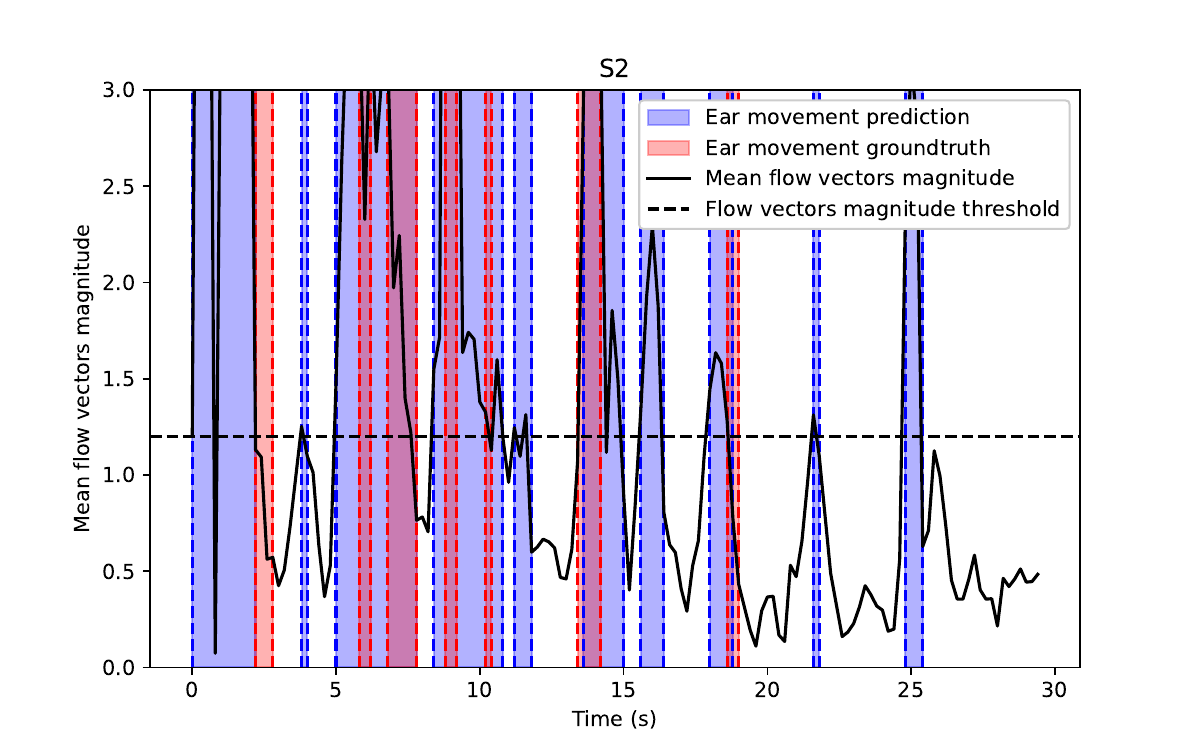}
        \caption{S2}
    \end{subfigure}
    
\end{figure}

\begin{figure}[H]
    \ContinuedFloat
    \captionsetup{labelformat=empty}
    \centering
    \begin{subfigure}{0.49\linewidth}
        \includegraphics[width=\linewidth]{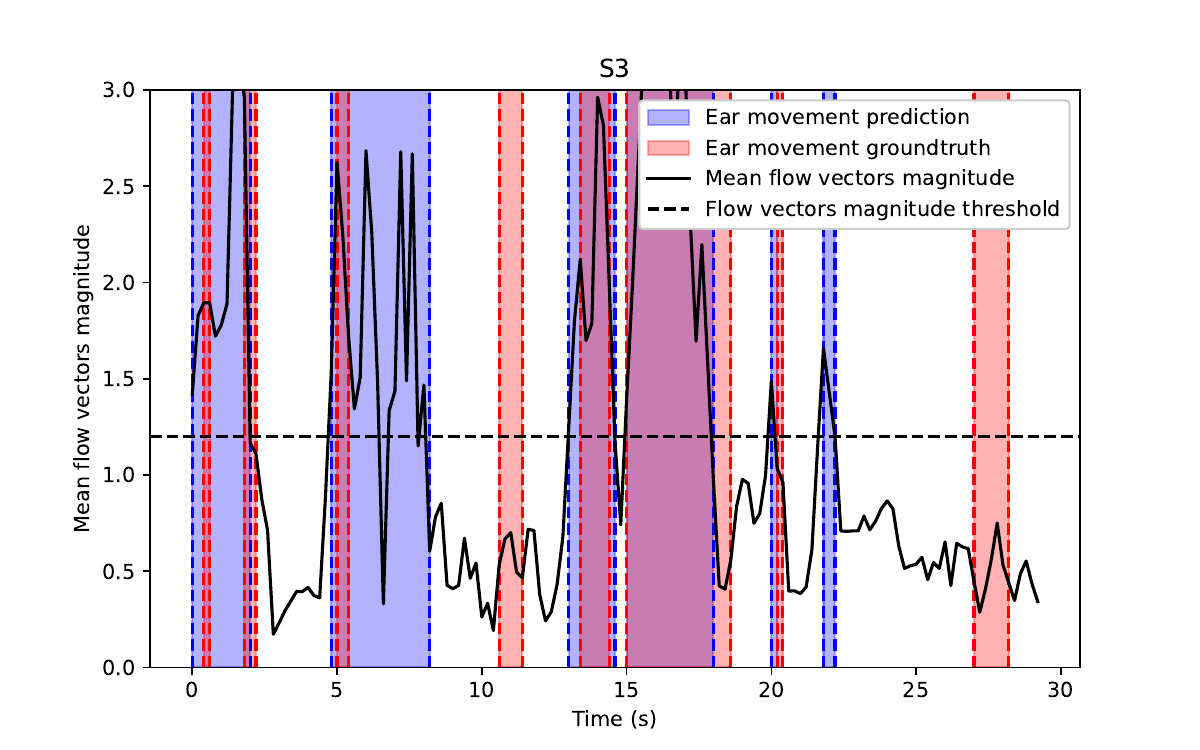}
        \caption{S3}
    \end{subfigure}
    \hfill
    \begin{subfigure}{0.49\linewidth}
        \includegraphics[width=\linewidth]{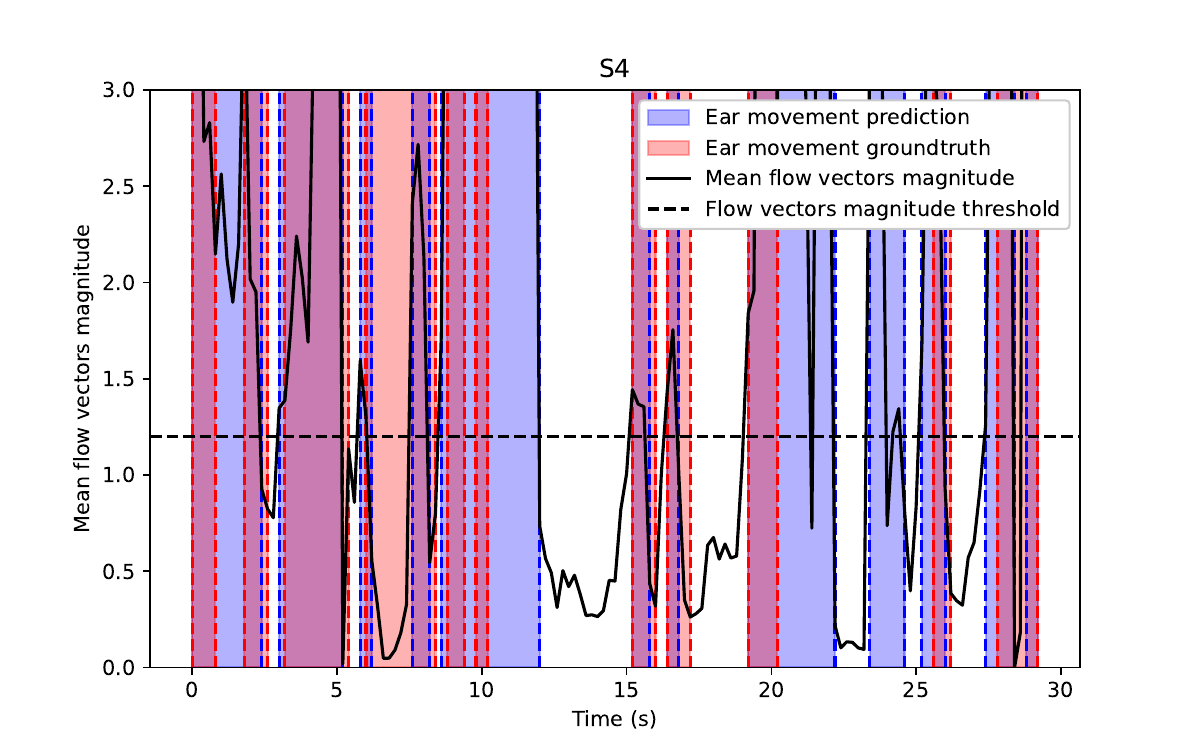}
        \caption{S4}
    \end{subfigure}
    
\end{figure}

\begin{figure}[H]
    \ContinuedFloat
    \captionsetup{labelformat=empty}
    \centering
    \begin{subfigure}{0.49\linewidth}
        \includegraphics[width=\linewidth]{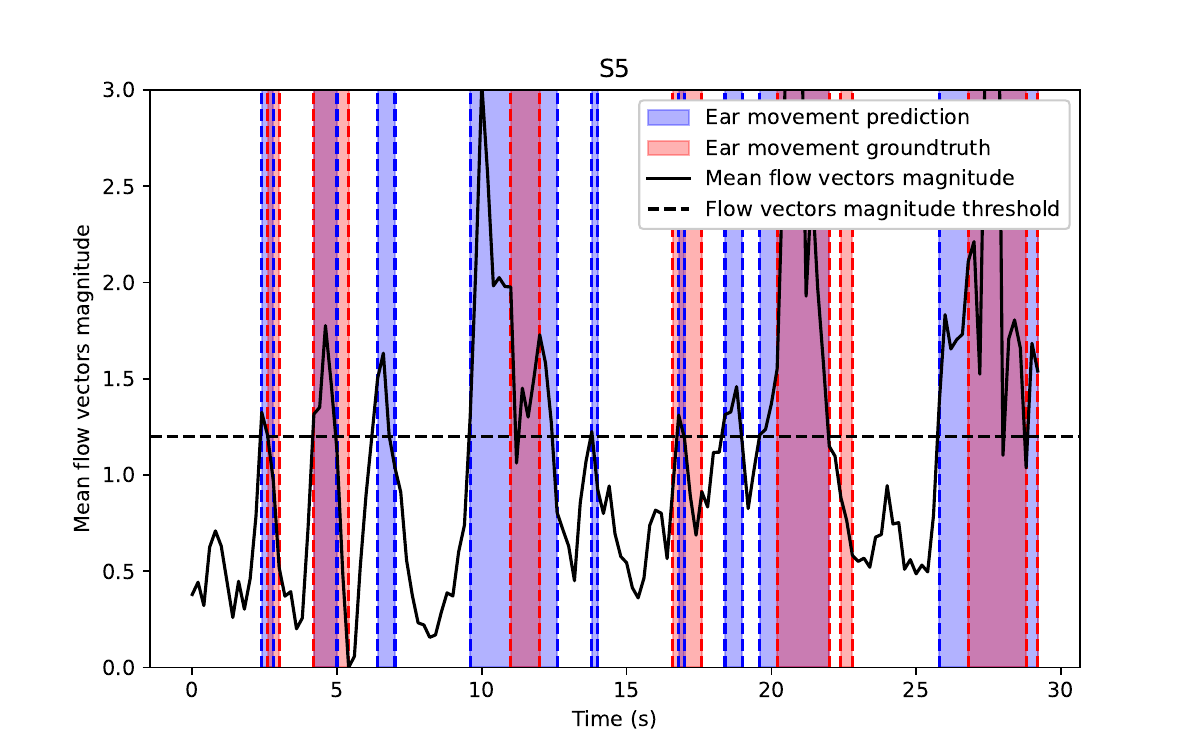}
        \caption{S5}
    \end{subfigure}
    \hfill
    \begin{subfigure}{0.49\linewidth}
        \includegraphics[width=\linewidth]{images/movDet_plots/S6.mp4.pdf}
        \caption{S6}
    \end{subfigure}
    
\end{figure}

\begin{figure}[H]
    \ContinuedFloat
    \captionsetup{labelformat=empty}
    \centering
    \begin{subfigure}{0.49\linewidth}
        \includegraphics[width=\linewidth]{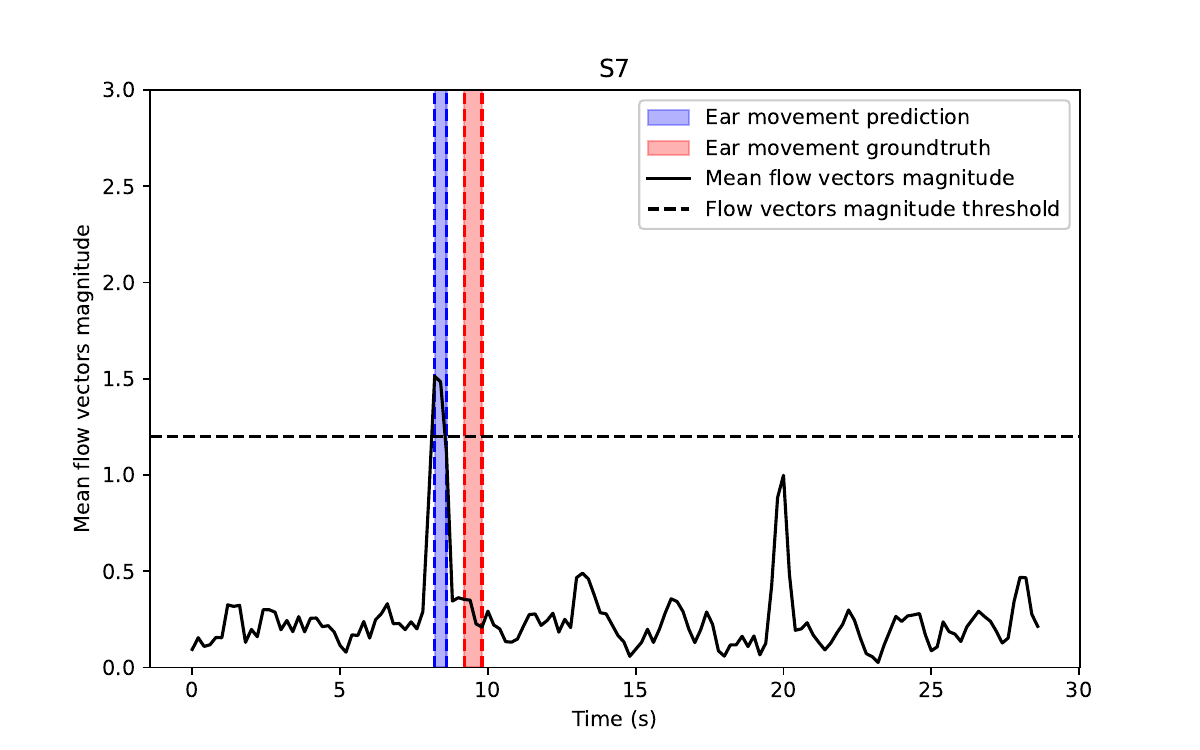}
        \caption{S7}
    \end{subfigure}
    \hfill
    \begin{subfigure}{0.49\linewidth}
        \includegraphics[width=\linewidth]{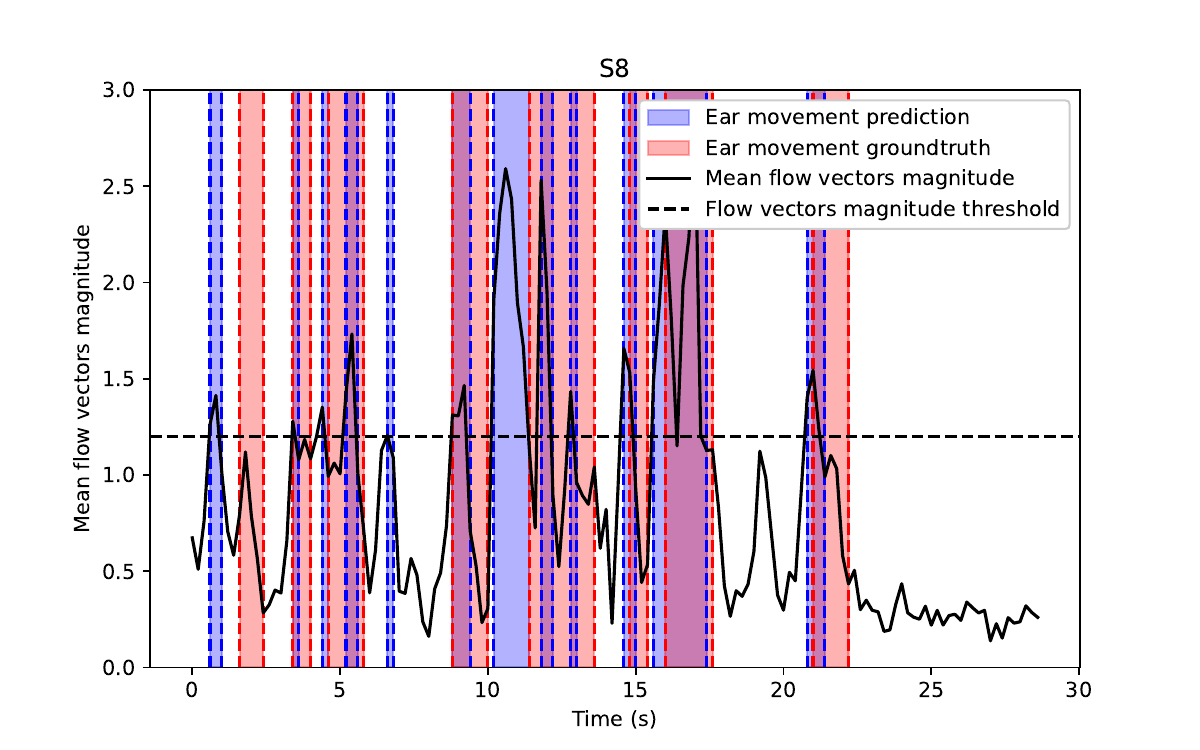}
        \caption{S8}
    \end{subfigure}
    
\end{figure}

\begin{figure}[H]
    \ContinuedFloat
    \captionsetup{labelformat=empty}
    \centering
    \begin{subfigure}{0.49\linewidth}
        \includegraphics[width=\linewidth]{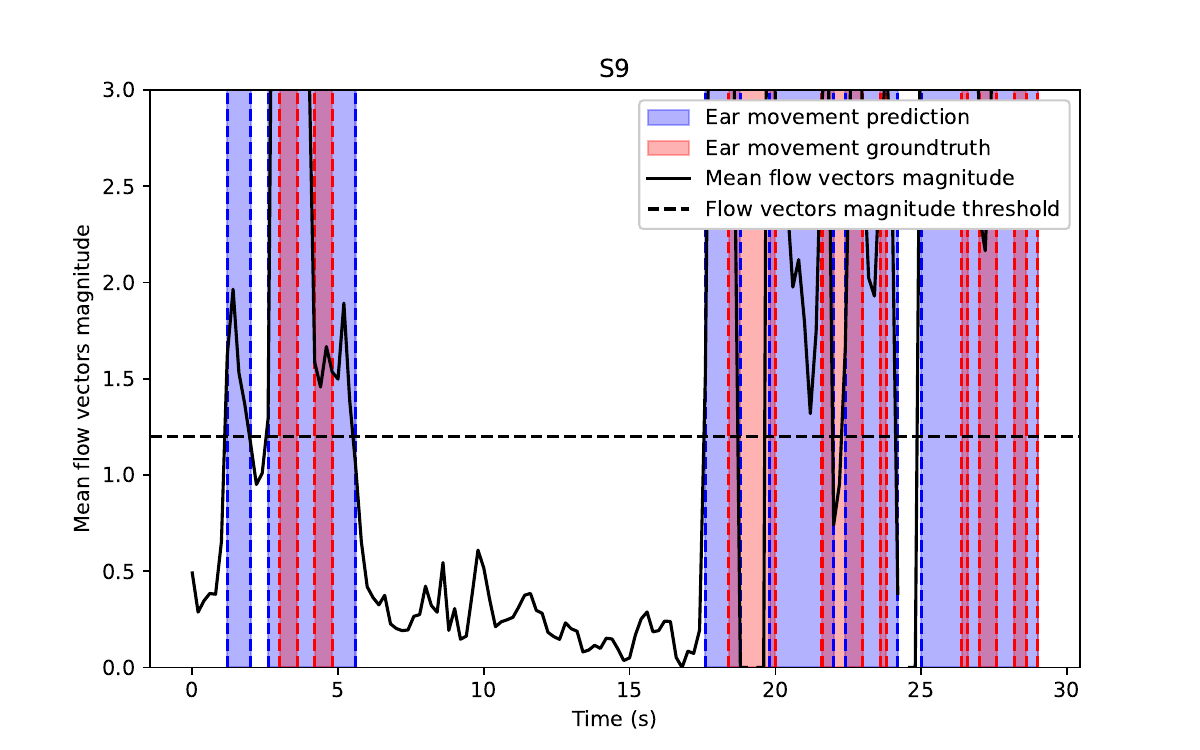}
        \caption{S9}
    \end{subfigure}
    \hfill
    \begin{subfigure}{0.49\linewidth}
        \includegraphics[width=\linewidth]{images/movDet_plots/S10.mp4.pdf}
        \caption{S10}
    \end{subfigure}
    
\end{figure}

\begin{figure}[H]
    \ContinuedFloat
    \captionsetup{labelformat=empty}
    \centering
    \begin{subfigure}{0.49\linewidth}
        \includegraphics[width=\linewidth]{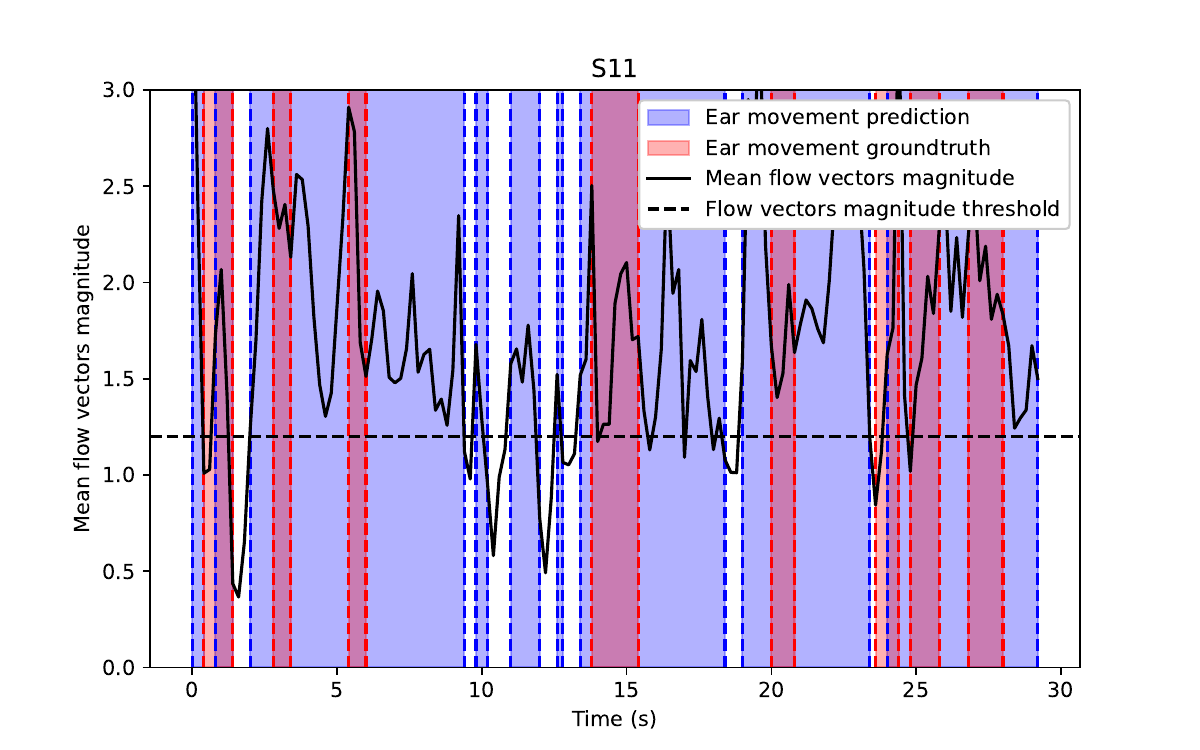}
        \caption{S11}
    \end{subfigure}
    \hfill
    \begin{subfigure}{0.49\linewidth}
        \includegraphics[width=\linewidth]{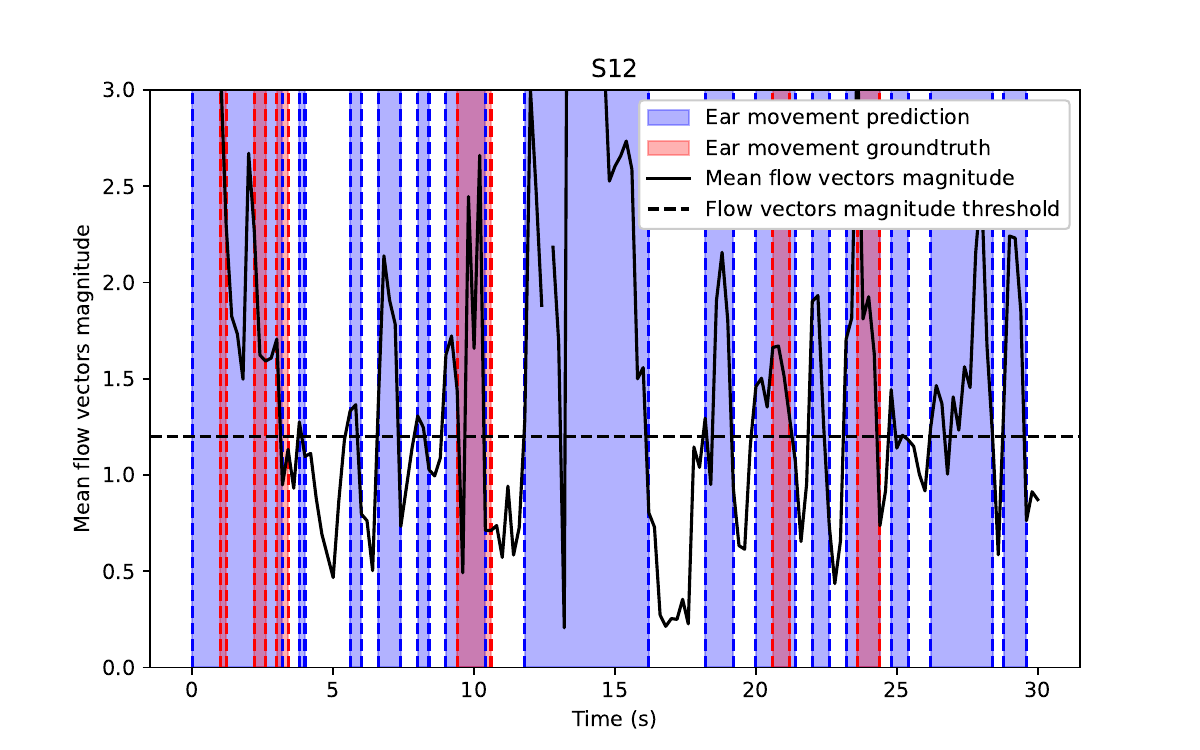}
        \caption{S12}
    \end{subfigure}
    \captionsetup{labelformat=default}
    \caption{Qualitative analysis for movDet method on full-length horse videos.}
    \label{fig:qual_supp_movdet}
\end{figure}

\section*{Qualitative Analysis: I3D+LSTM}

\begin{figure}[H]
    \captionsetup{labelformat=empty}
    \centering
    \begin{subfigure}{0.49\linewidth}
        \includegraphics[width=\linewidth]{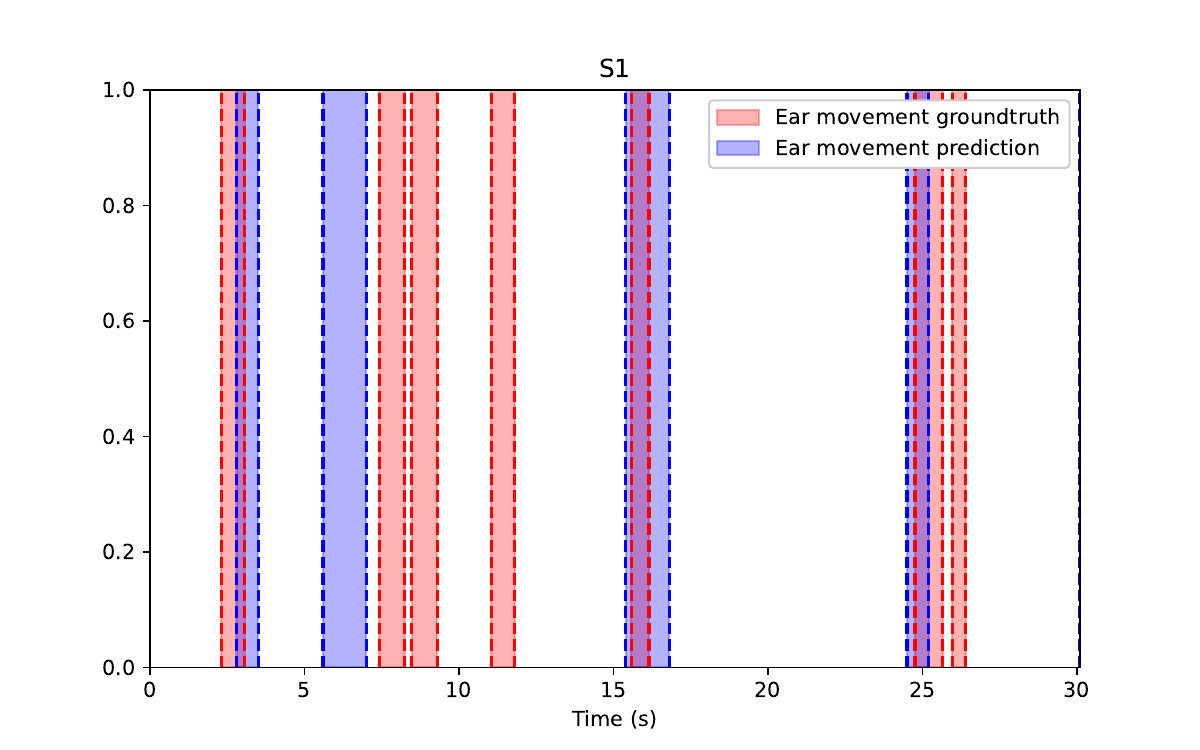}
        \caption{S1}
    \end{subfigure}
    \hfill
    \begin{subfigure}{0.49\linewidth}
        \includegraphics[width=\linewidth]{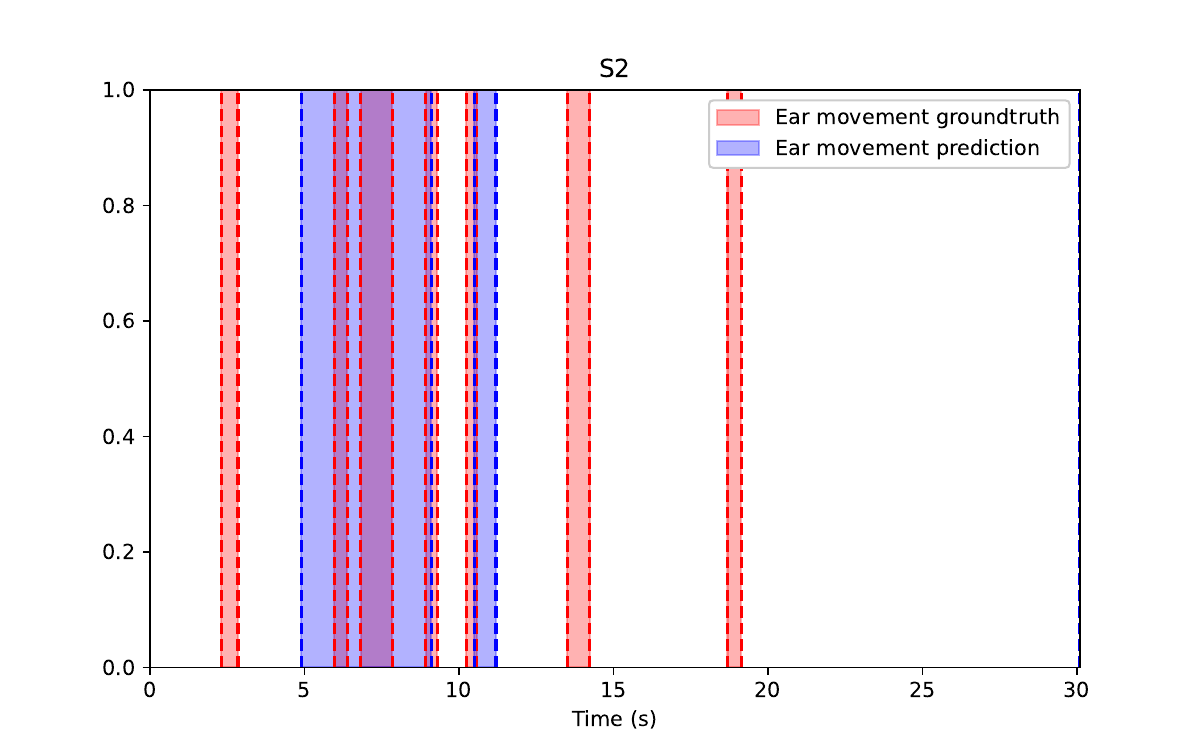}
        \caption{S2}
    \end{subfigure}
    
\end{figure}

\begin{figure}[H]
    \ContinuedFloat
    \captionsetup{labelformat=empty}
    \centering
    \begin{subfigure}{0.49\linewidth}
        \includegraphics[width=\linewidth]{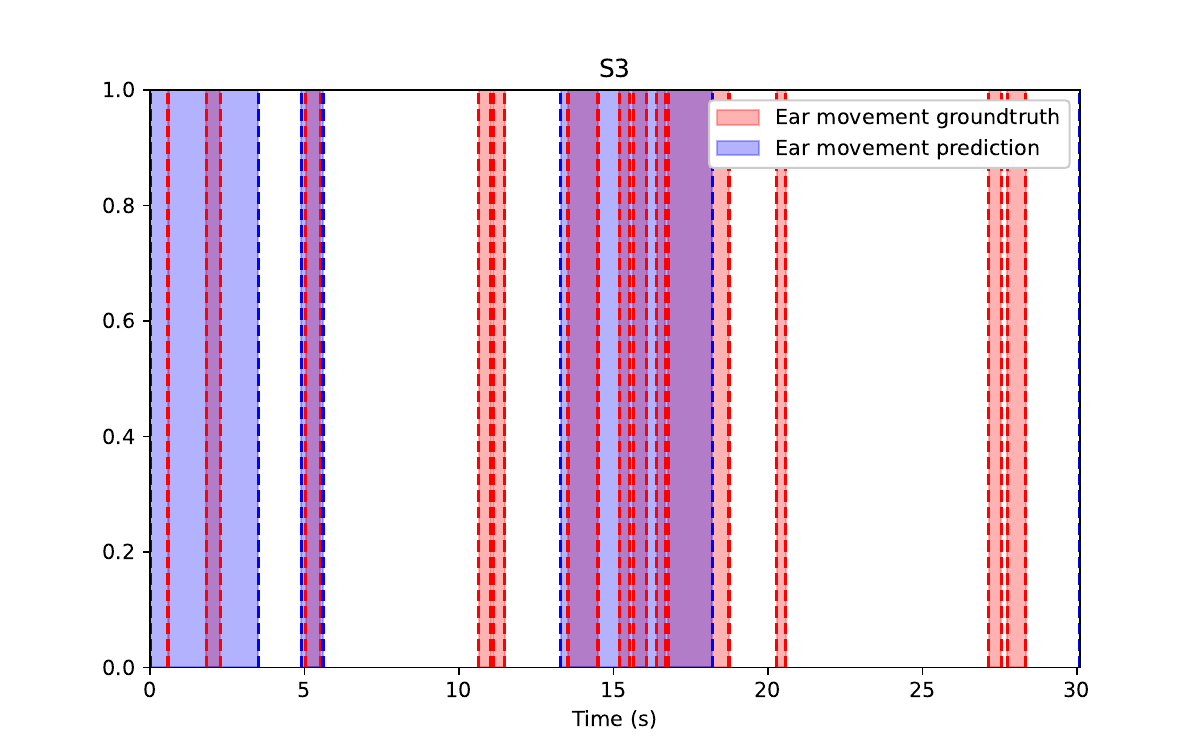}
        \caption{S3}
    \end{subfigure}
    \hfill
    \begin{subfigure}{0.49\linewidth}
        \includegraphics[width=\linewidth]{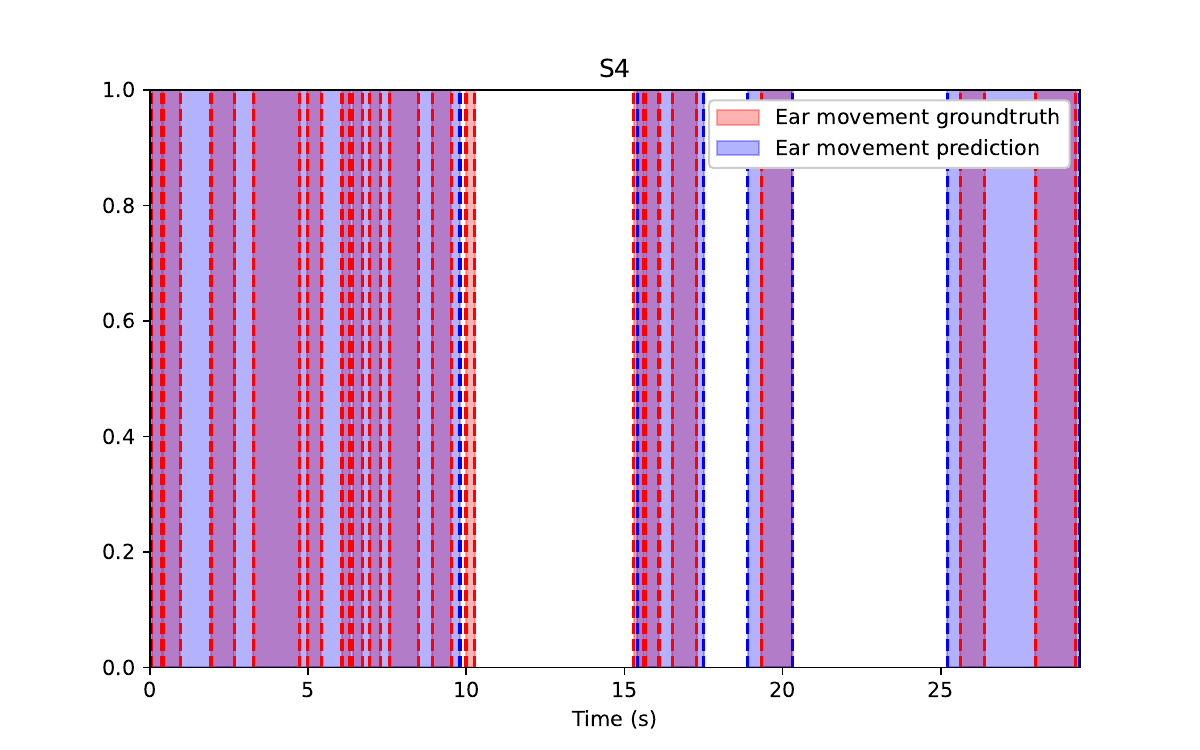}
        \caption{S4}
    \end{subfigure}

\end{figure}

\begin{figure}[H]
    \ContinuedFloat
    \captionsetup{labelformat=empty}
    \centering
    \begin{subfigure}{0.49\linewidth}
        \includegraphics[width=\linewidth]{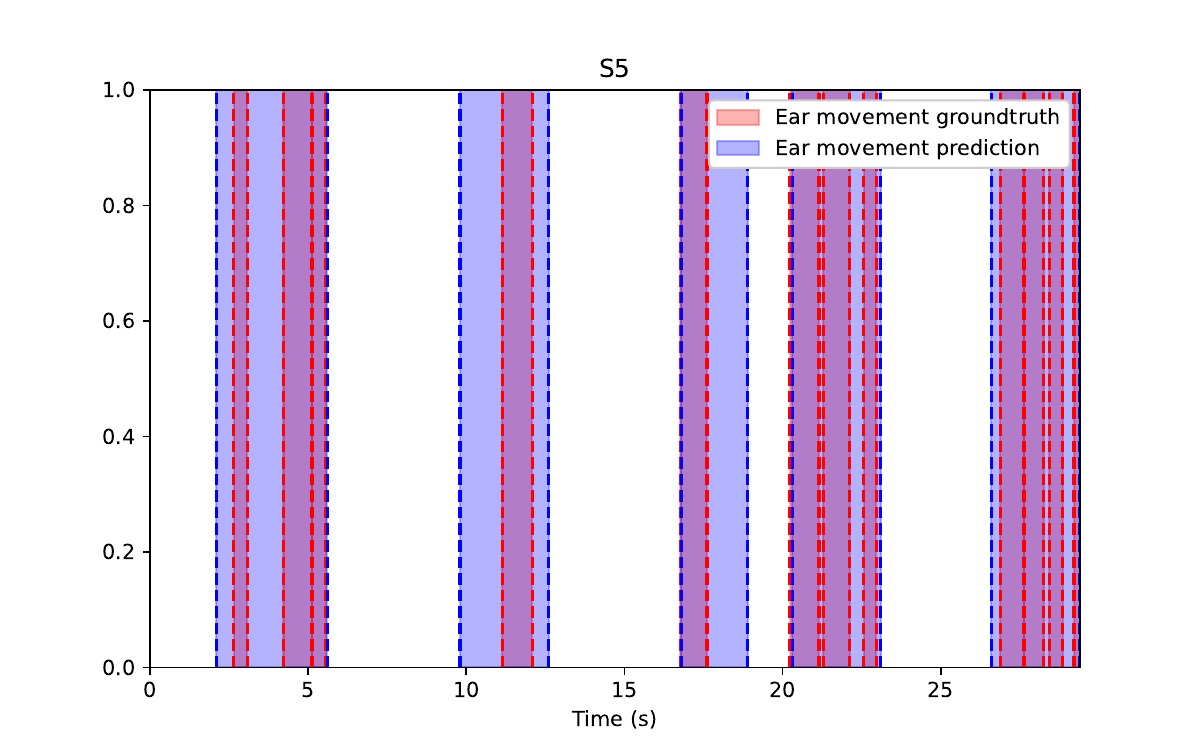}
        \caption{S5}
    \end{subfigure}
    \hfill
    \begin{subfigure}{0.49\linewidth}
        \includegraphics[width=\linewidth]{images/i3dLSTM_plots/S6.mp4.pdf}
        \caption{S6}
    \end{subfigure}
    
\end{figure}

\begin{figure}[H]
    \ContinuedFloat
    \captionsetup{labelformat=empty}
    \centering
    \begin{subfigure}{0.49\linewidth}
        \includegraphics[width=\linewidth]{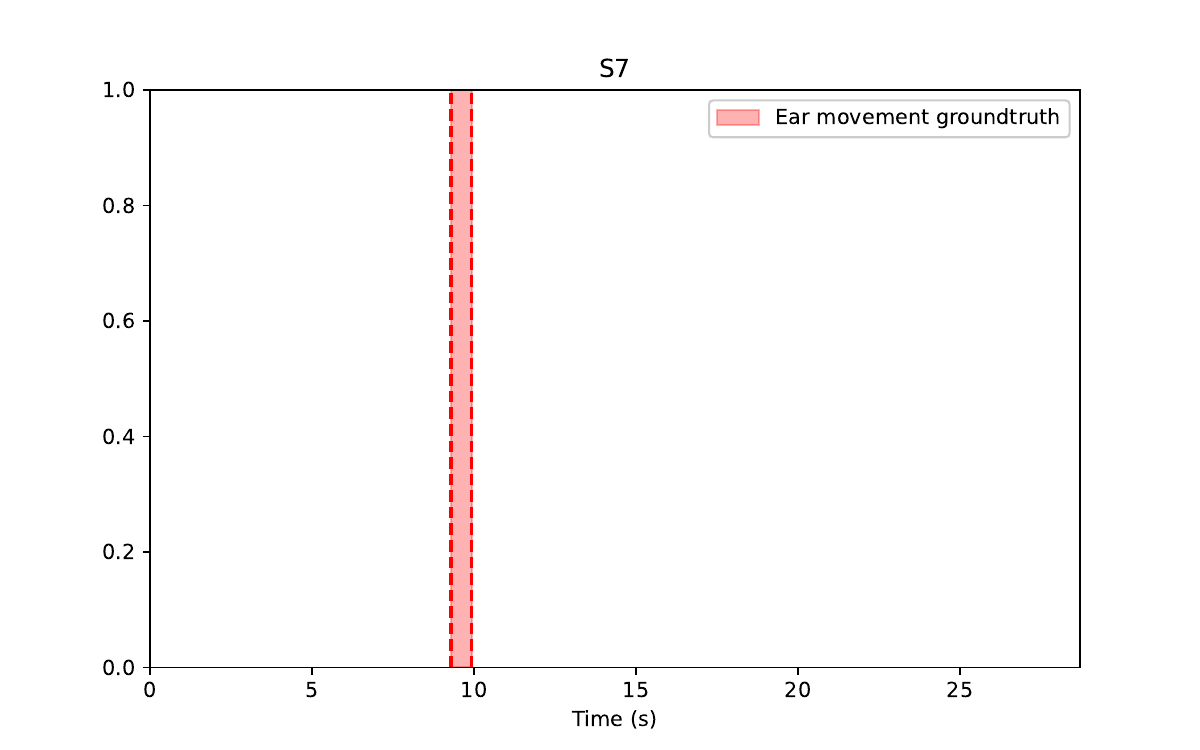}
        \caption{S7}
    \end{subfigure}
    \hfill
    \begin{subfigure}{0.49\linewidth}
        \includegraphics[width=\linewidth]{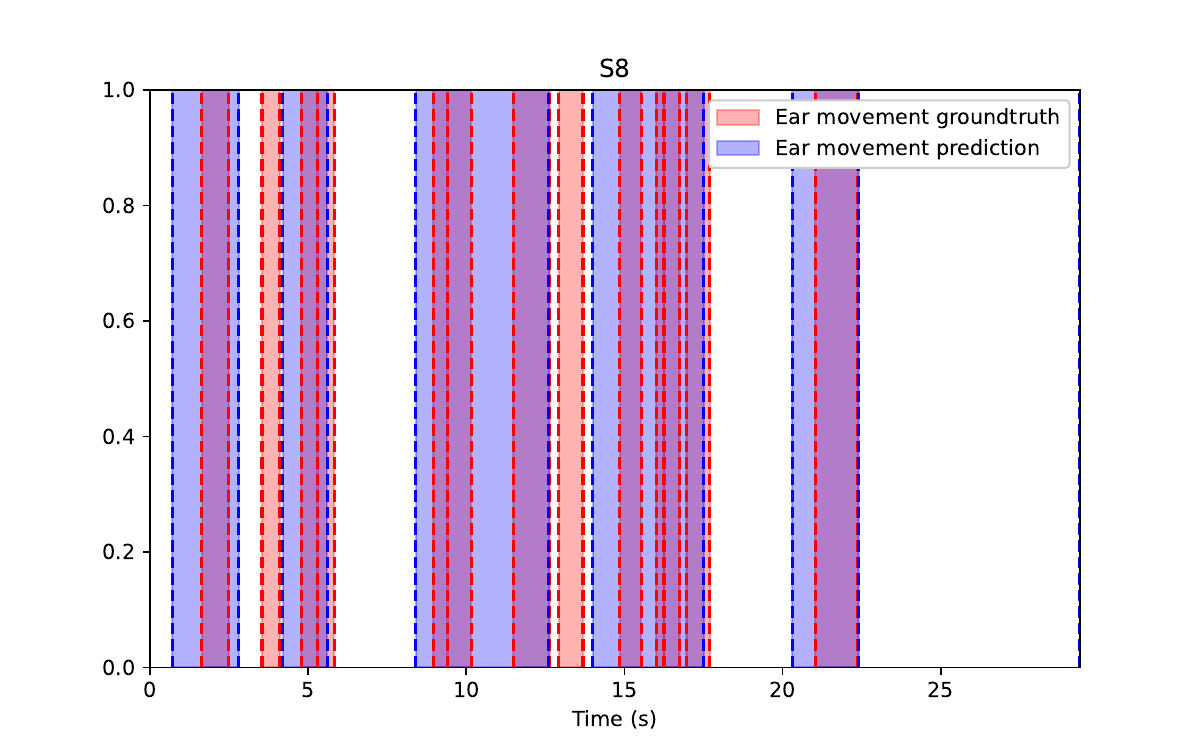}
        \caption{S8}
    \end{subfigure}
    
\end{figure}

\begin{figure}[H]
    \ContinuedFloat
    \captionsetup{labelformat=empty}
    \centering
    \begin{subfigure}{0.49\linewidth}
        \includegraphics[width=\linewidth]{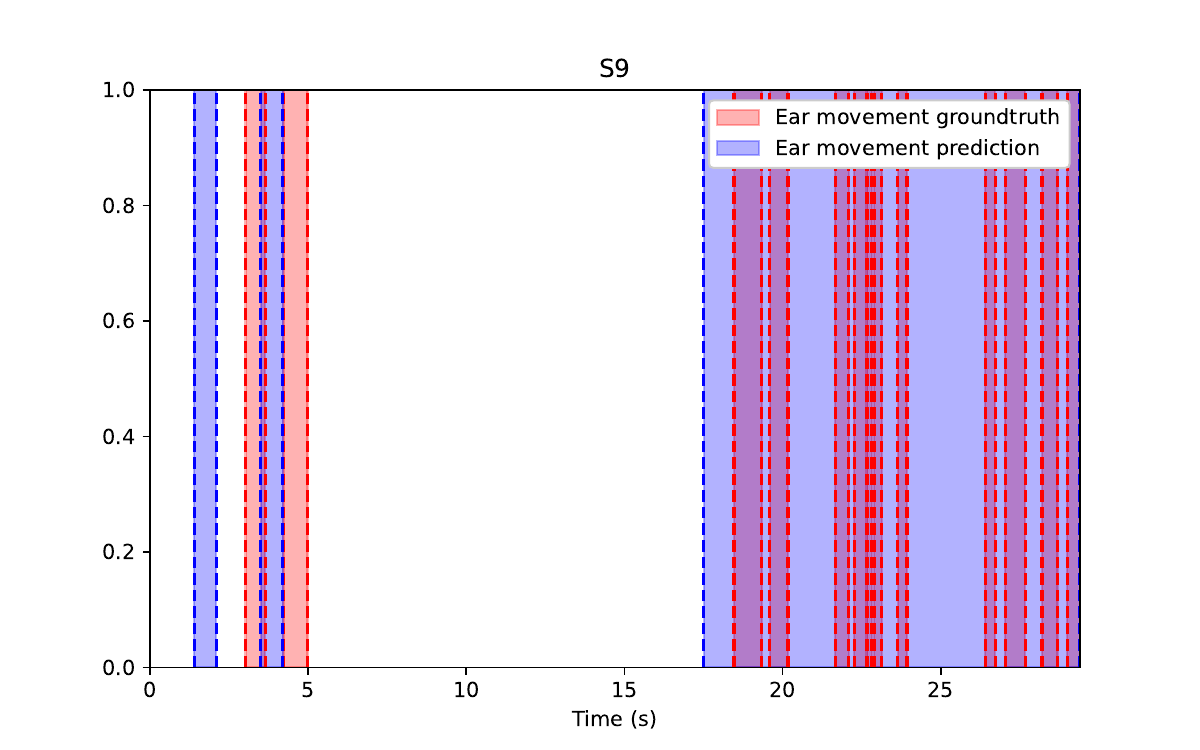}
        \caption{S9}
    \end{subfigure}
    \hfill
    \begin{subfigure}{0.49\linewidth}
        \includegraphics[width=\linewidth]{images/i3dLSTM_plots/S10.mp4.pdf}
        \caption{S10}
    \end{subfigure}
    
\end{figure}

\begin{figure}[H]
    \ContinuedFloat
    \captionsetup{labelformat=empty}
    \centering
    \begin{subfigure}{0.49\linewidth}
        \includegraphics[width=\linewidth]{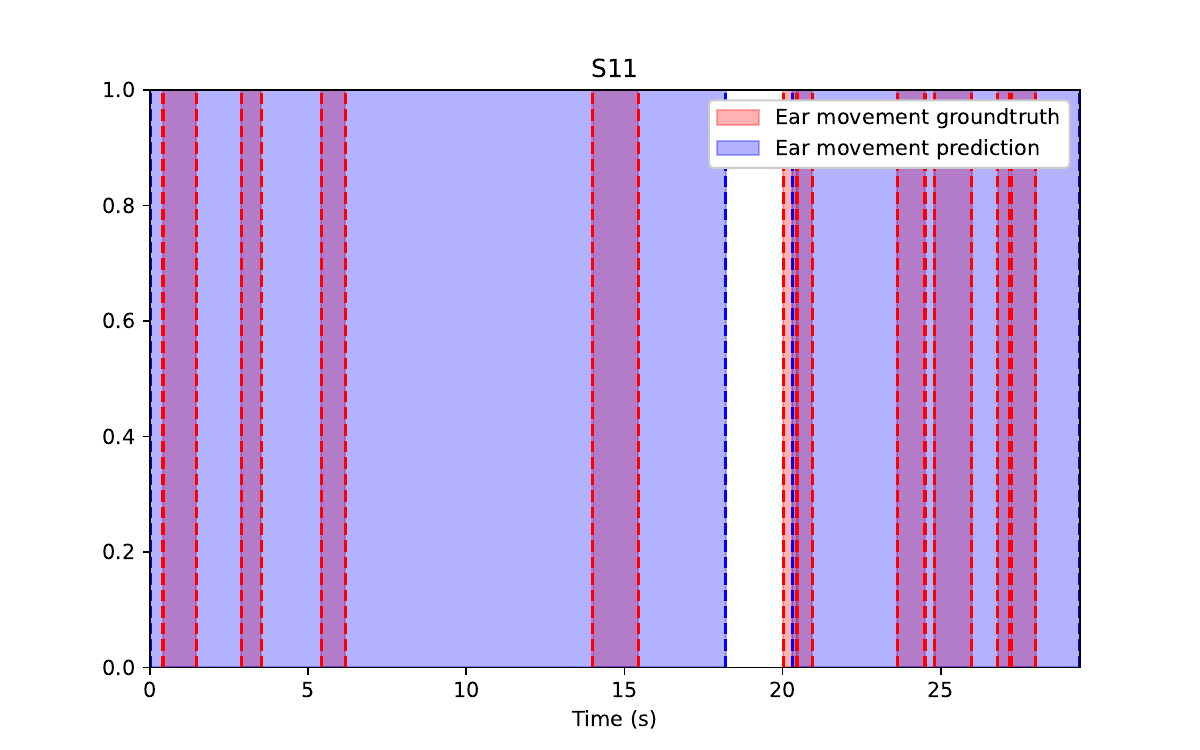}
        \caption{S11}
    \end{subfigure}
    \hfill
    \begin{subfigure}{0.49\linewidth}
        \includegraphics[width=\linewidth]{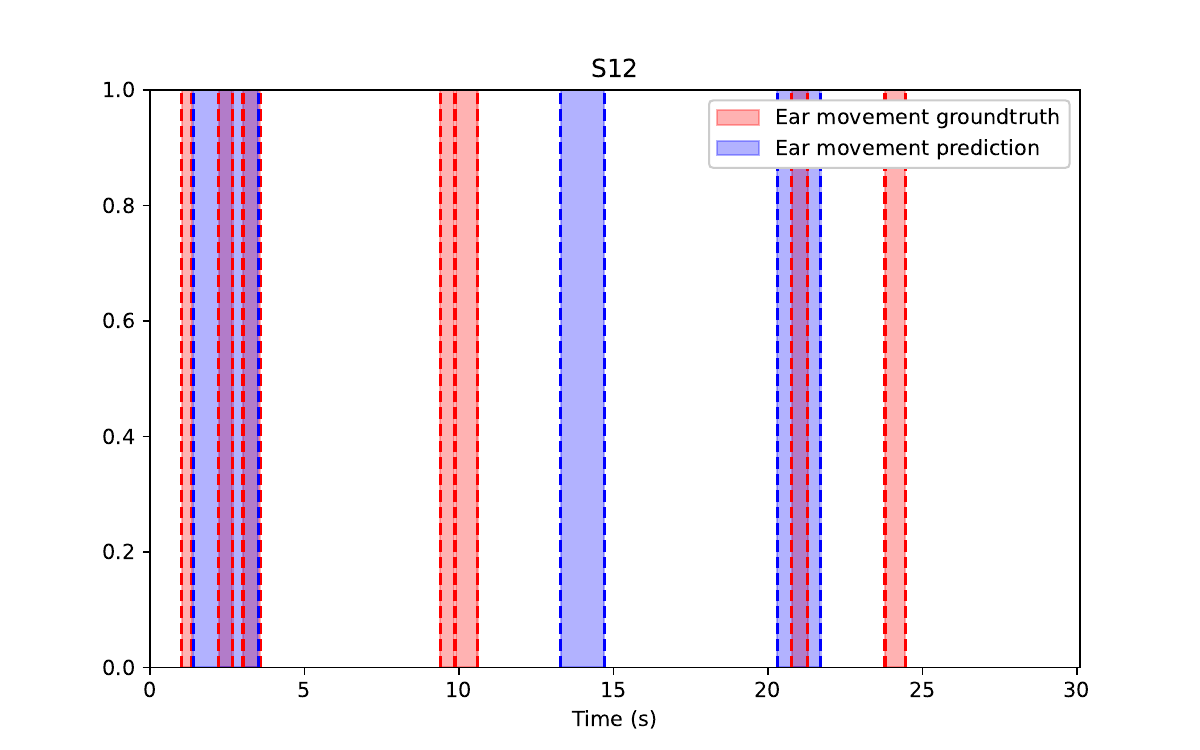}
        \caption{S12}
    \end{subfigure}
    \captionsetup{labelformat=default}
    \caption{Qualitative analysis for I3D+LSTM method on full-length horse videos.}
    \label{fig:qual_supp_i3d_lstm}
\end{figure}

\section*{Qualitative Analysis: VideoMAE+LSTM}

\begin{figure}[H]
    \captionsetup{labelformat=empty}
    \centering
    \begin{subfigure}{0.49\linewidth}
        \includegraphics[width=\linewidth]{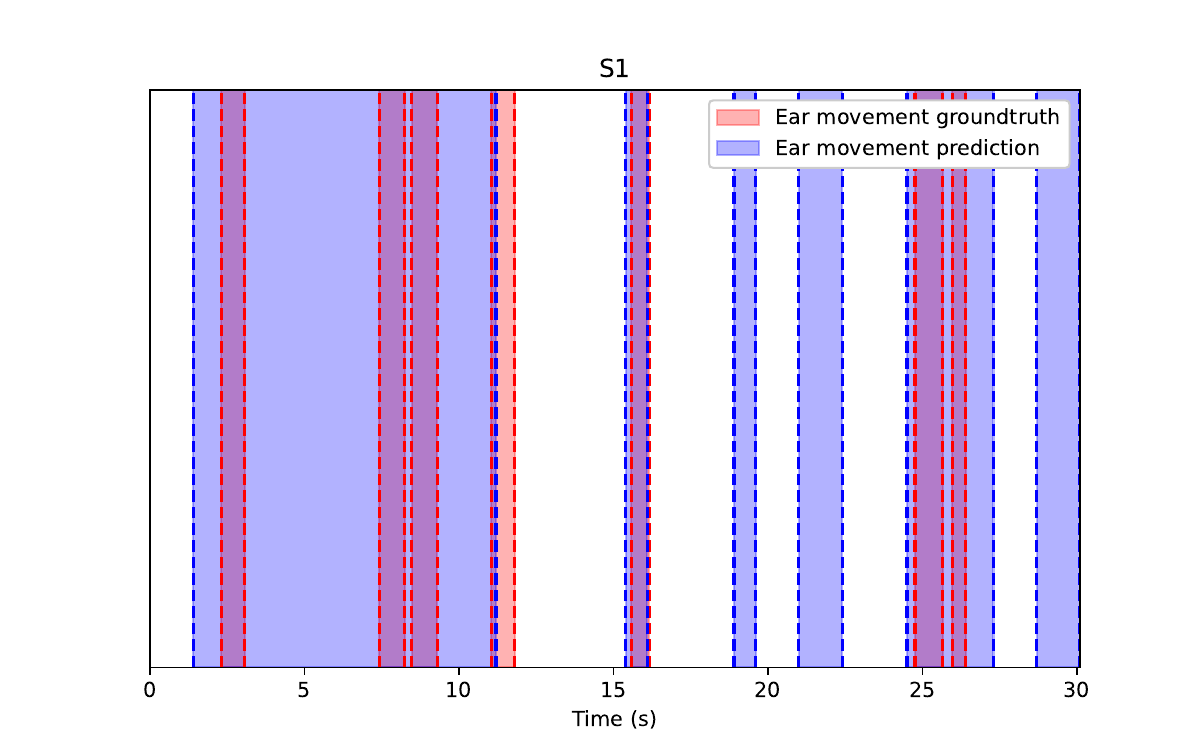}
        \caption{S1}
    \end{subfigure}
    \hfill
    \begin{subfigure}{0.49\linewidth}
        \includegraphics[width=\linewidth]{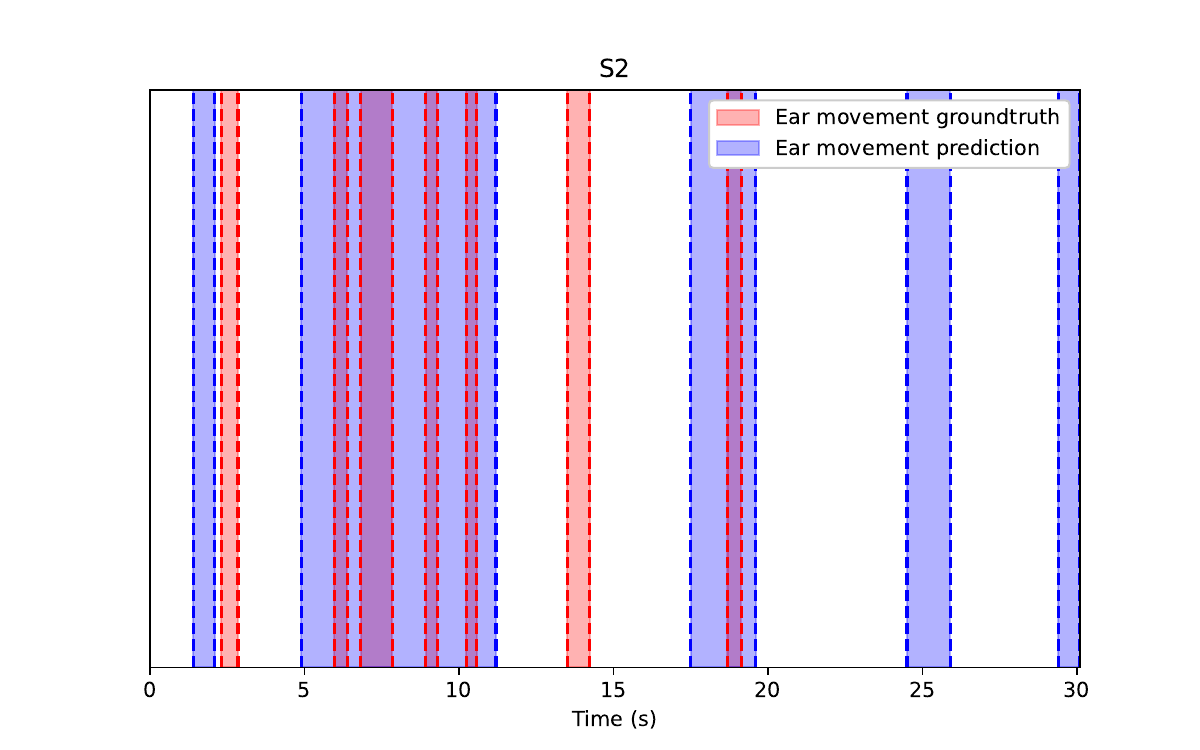}
        \caption{S2}
    \end{subfigure}
    
\end{figure}

\begin{figure}[H]
    \ContinuedFloat
    \captionsetup{labelformat=empty}
    \centering
    \begin{subfigure}{0.49\linewidth}
        \includegraphics[width=\linewidth]{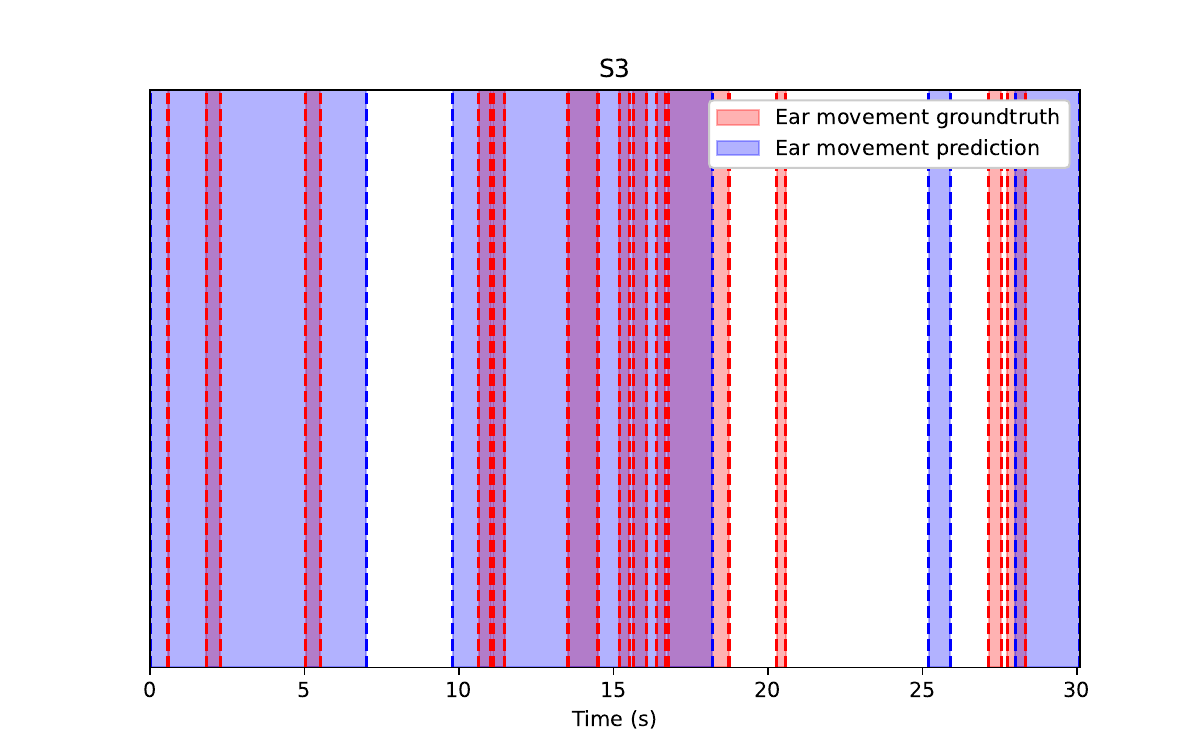}
        \caption{S3}
    \end{subfigure}
    \hfill
    \begin{subfigure}{0.49\linewidth}
        \includegraphics[width=\linewidth]{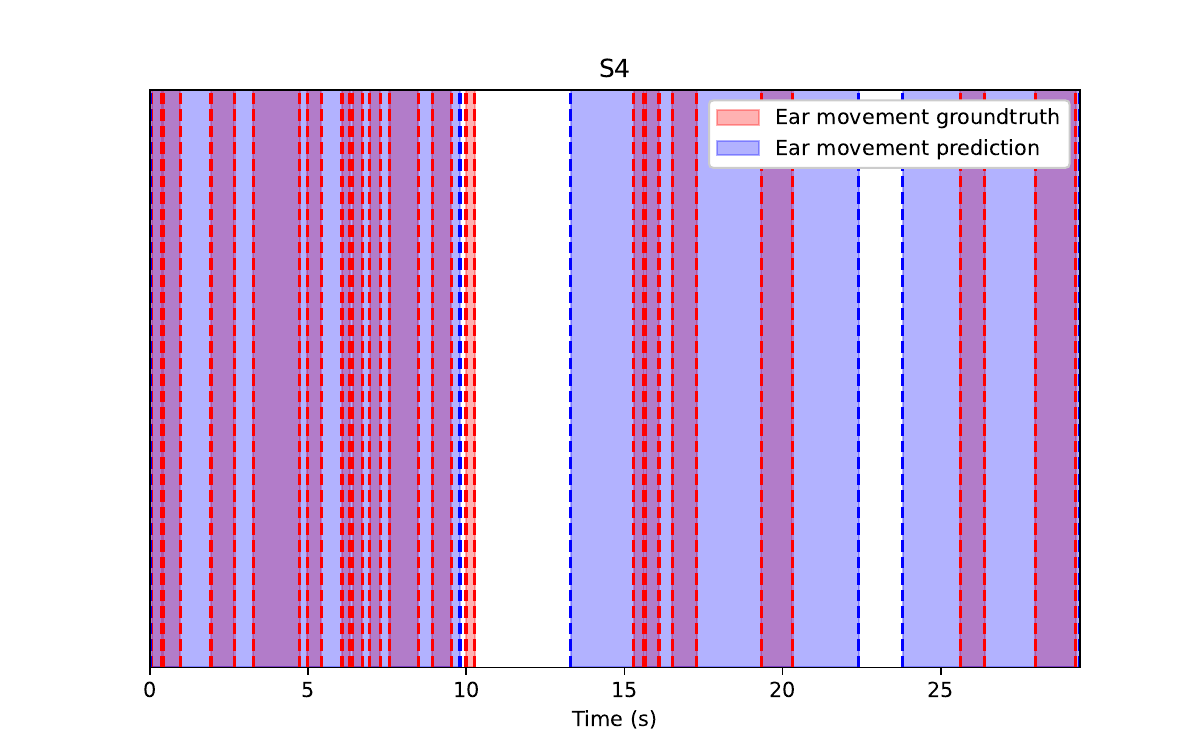}
        \caption{S4}
    \end{subfigure}
    
\end{figure}

\begin{figure}[H]
    \ContinuedFloat
    \captionsetup{labelformat=empty}
    \centering
    \begin{subfigure}{0.49\linewidth}
        \includegraphics[width=\linewidth]{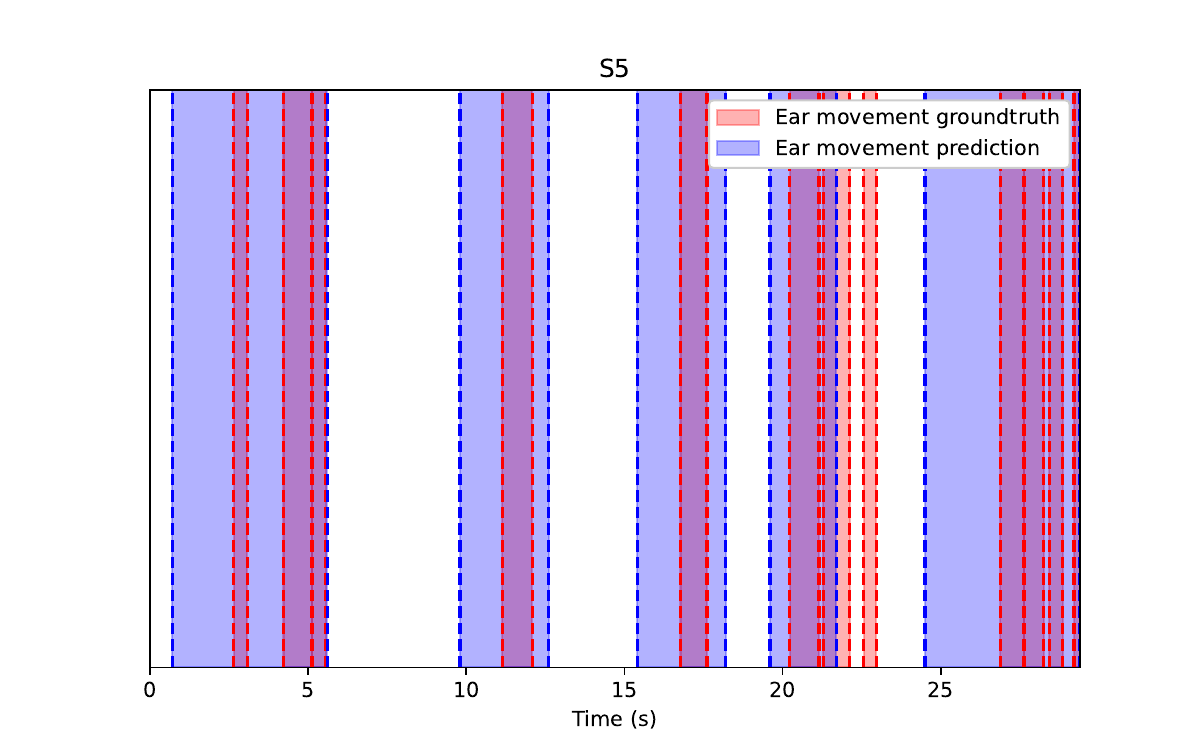}
        \caption{S5}
    \end{subfigure}
    \hfill
    \begin{subfigure}{0.49\linewidth}
        \includegraphics[width=\linewidth]{images/videoMAELSTM_plots/S6.mp4.pdf}
        \caption{S6}
    \end{subfigure}
    
\end{figure}

\begin{figure}[H]
    \ContinuedFloat
    \captionsetup{labelformat=empty}
    \centering
    \begin{subfigure}{0.49\linewidth}
        \includegraphics[width=\linewidth]{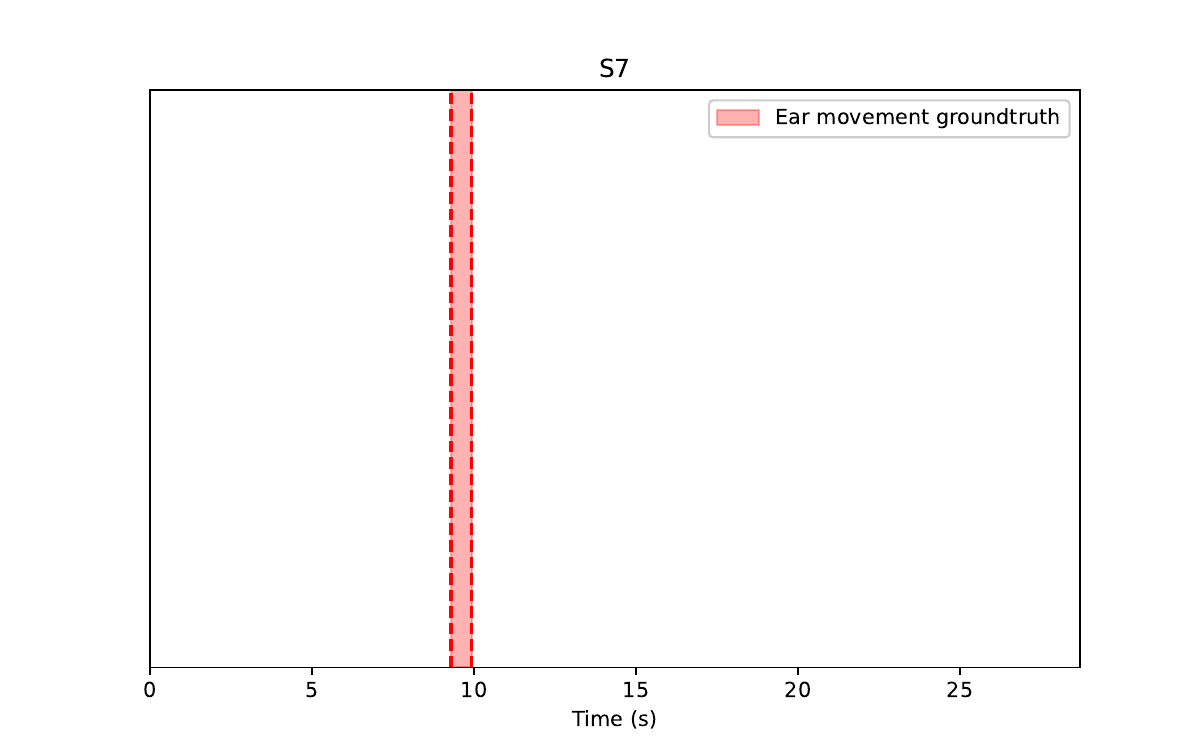}
        \caption{S7}
    \end{subfigure}
    \hfill
    \begin{subfigure}{0.49\linewidth}
        \includegraphics[width=\linewidth]{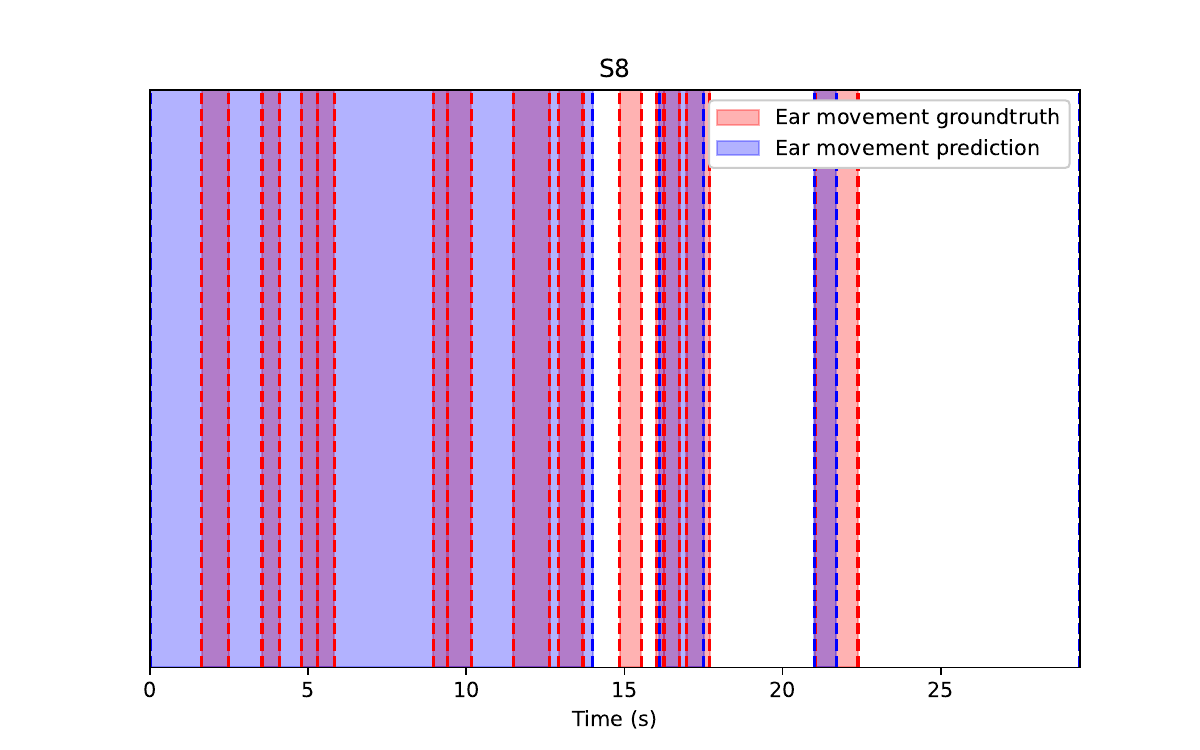}
        \caption{S8}
    \end{subfigure}
    
\end{figure}

\begin{figure}[H]
    \ContinuedFloat
    \captionsetup{labelformat=empty}
    \centering
    \begin{subfigure}{0.49\linewidth}
        \includegraphics[width=\linewidth]{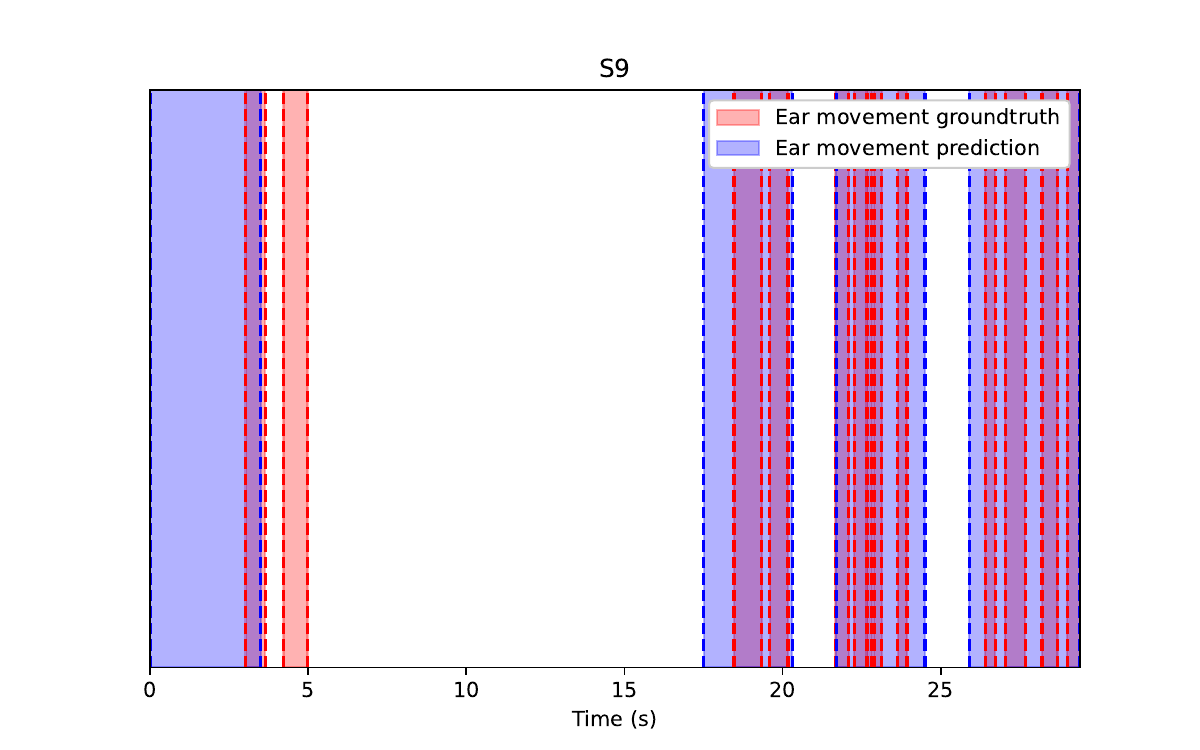}
        \caption{S9}
    \end{subfigure}
    \hfill
    \begin{subfigure}{0.49\linewidth}
        \includegraphics[width=\linewidth]{images/videoMAELSTM_plots/S10.mp4.pdf}
        \caption{S10}
    \end{subfigure}
    
\end{figure}

\begin{figure}[H]
    \ContinuedFloat
    \captionsetup{labelformat=empty}
    \centering
    \begin{subfigure}{0.49\linewidth}
        \includegraphics[width=\linewidth]{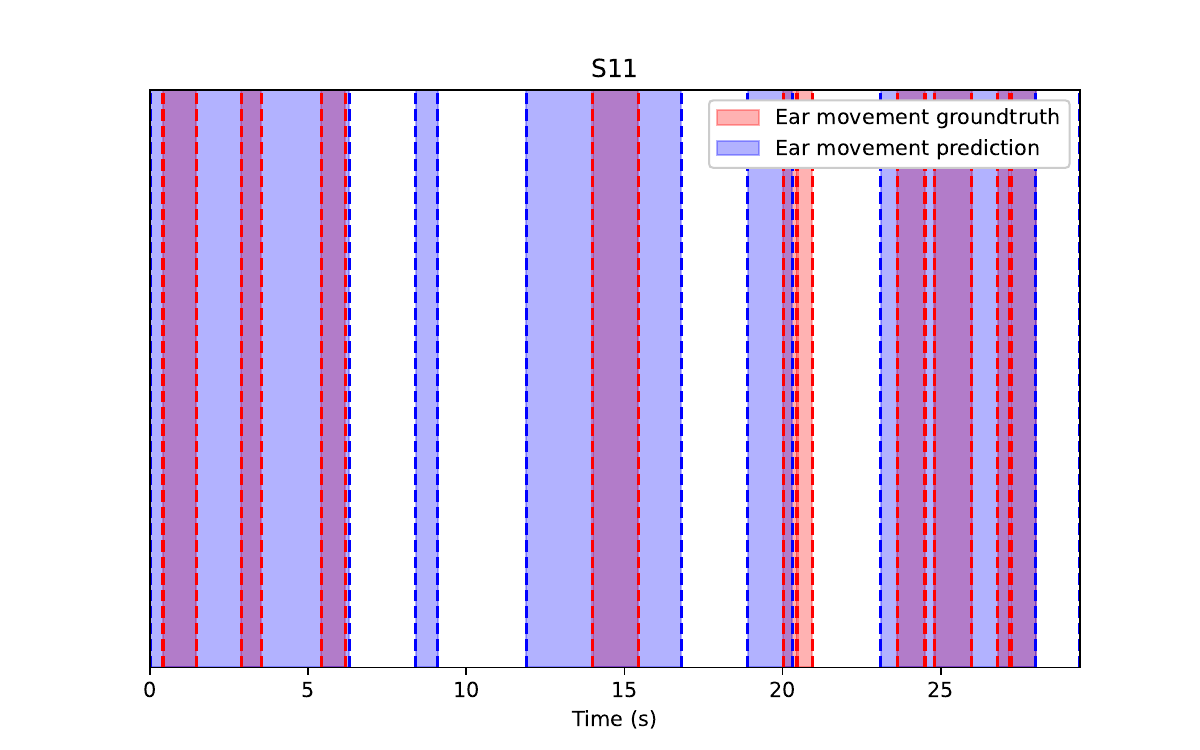}
        \caption{S11}
    \end{subfigure}
    \hfill
    \begin{subfigure}{0.49\linewidth}
        \includegraphics[width=\linewidth]{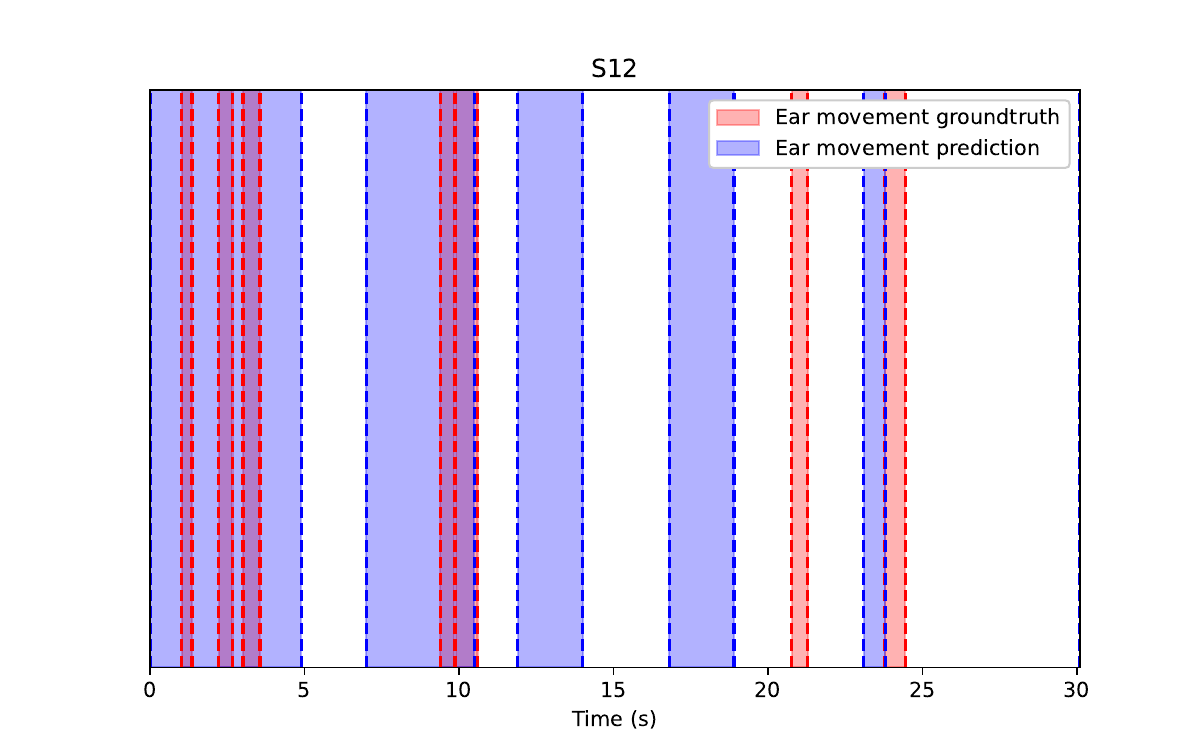}
        \caption{S12}
    \end{subfigure}
    \captionsetup{labelformat=default}
    \caption{Qualitative analysis for VideoMAE+LSTM method on full-length horse videos.}
    \label{fig:qual_supp_videomae_lstm}
\end{figure}


\end{document}